\documentclass[letterpaper]{article} % DO NOT CHANGE THIS
\usepackage{aaai2026}  % DO NOT CHANGE THIS
\usepackage{times}  % DO NOT CHANGE THIS
\usepackage{helvet}  % DO NOT CHANGE THIS
\usepackage{courier}  % DO NOT CHANGE THIS
\usepackage[hyphens]{url}  % DO NOT CHANGE THIS
\usepackage{graphicx} % DO NOT CHANGE THIS
\usepackage{natbib}  % DO NOT CHANGE THIS AND DO NOT ADD ANY OPTIONS TO IT
\usepackage{caption} % DO NOT CHANGE THIS AND DO NOT ADD ANY OPTIONS TO IT
\usepackage{algorithm}
\usepackage{algorithmic}

\usepackage{amssymb}
\usepackage{algorithm}
\usepackage{algorithmic}

\usepackage{multirow}
\usepackage{pifont}

\usepackage{xcolor}

\usepackage{amsmath}
\usepackage{amssymb}
\usepackage{mathtools}
\usepackage{booktabs}
\usepackage{newfloat}
\usepackage{listings}
\DeclareCaptionStyle{ruled}{labelfont=normalfont,labelsep=colon,strut=off} % DO NOT CHANGE THIS
\floatstyle{ruled}
\newfloat{listing}{tb}{lst}{}
\floatname{listing}{Listing}
\def\eg{\emph{e.g.,~}}
\def\ie{\emph{i.e.,~}}
\def\ournet{DriveFlow}

\title{DriveFlow: Rectified Flow Adaptation for \\ Robust 3D Object Detection in Autonomous Driving}
\author{
    Hongbin Lin \textsuperscript{\rm 1,2},
    Yiming Yang \textsuperscript{\rm 1,2}, 
    Chaoda Zheng \textsuperscript{\rm 3}, 
    Yifan Zhang \textsuperscript{\rm 4}, \\
    Shuaicheng Niu \textsuperscript{\rm 5}, 
    Zilu Guo \textsuperscript{\rm 1,2}, 
    Yafeng Li \textsuperscript{\rm 6}, 
    Gui Gui \textsuperscript{\rm 7}, 
    Shuguang Cui \textsuperscript{\rm 2,1}, 
    Zhen Li \textsuperscript{\rm 2,1} \thanks{Corresponding author.}
}
\affiliations{
    \textsuperscript{\rm 1} FNii-Shenzhen,
    \textsuperscript{\rm 2} SSE, CUHK-Shenzhen,
    \textsuperscript{\rm 3} Xpeng Motors,
    \textsuperscript{\rm 4} MiroMind AI, \\
    \textsuperscript{\rm 5} Nanyang Technological University,
    \textsuperscript{\rm 6} Baoji University of Arts and Sciences,
    \textsuperscript{\rm 7} Central South University, 
}

\usepackage{bibentry}
\begin{document}

\maketitle

\begin{abstract}
In autonomous driving, vision-centric 3D object detection recognizes and localizes 3D objects from RGB images. However, due to high annotation costs and diverse outdoor scenes, training data often fails to cover all possible test scenarios, known as the out-of-distribution (OOD) issue.
Training-free image editing offers a promising solution for improving model robustness by training data enhancement without any modifications to pre-trained diffusion models.
Nevertheless, inversion-based methods often suffer from limited effectiveness and inherent inaccuracies, while recent rectified-flow-based approaches struggle to preserve objects with accurate 3D geometry.
In this paper, we propose \emph{\textbf{\ournet}}, a Rectified Flow Adaptation method for training data enhancement in autonomous driving based on pre-trained Text-to-Image flow models.
Based on frequency decomposition, \textbf{\ournet} introduces two strategies to adapt noise-free editing paths derived from text-conditioned velocities.
1) High-Frequency Foreground Preservation: \ournet~incorporates a high-frequency alignment loss for foreground to maintain precise 3D object geometry.
2) Dual-Frequency Background Optimization: \ournet~also conducts dual-frequency optimization for background, balancing editing flexibility and semantic consistency.
Extensive experiments validate the effectiveness and efficiency of \ournet, demonstrating comprehensive performance improvements across OOD scenarios.
\end{abstract}

\begin{links}
    \link{Code}{https://github.com/Hongbin98/DriveFlow}
    % \link{Datasets}{https://aaai.org/example/datasets}
    % \link{Extended version}{https://aaai.org/example/extended-version}
\end{links}

\section{Introduction}
Three-dimensional (3D) Object Detection constitutes a critical computer vision challenge, involving the identification and localization of objects within three-dimensional space using various sensing modalities~\cite{li2022bevformer,chen2023voxelnext}. 
Due to the economic advantages, vision-centric 3D detection has emerged as a prominent paradigm that leverages solely RGB images from single or multiple cameras, complemented by calibration information~\cite{xu2023mononerd,wang2023exploring,yan2024monocd,pu2025monodgp}. 
Given the inherent challenges in vision-centric detection, existing methods~\cite{oh2025monowad,lin2025monotta,zhang2025geobev,li2025rctrans} have still achieved remarkable progress over various benchmarks~\cite{geiger2012we,caesar2020nuscenes,sun2020scalability}.

\begin{figure}[t]
\centering
\includegraphics[width=0.9\linewidth]{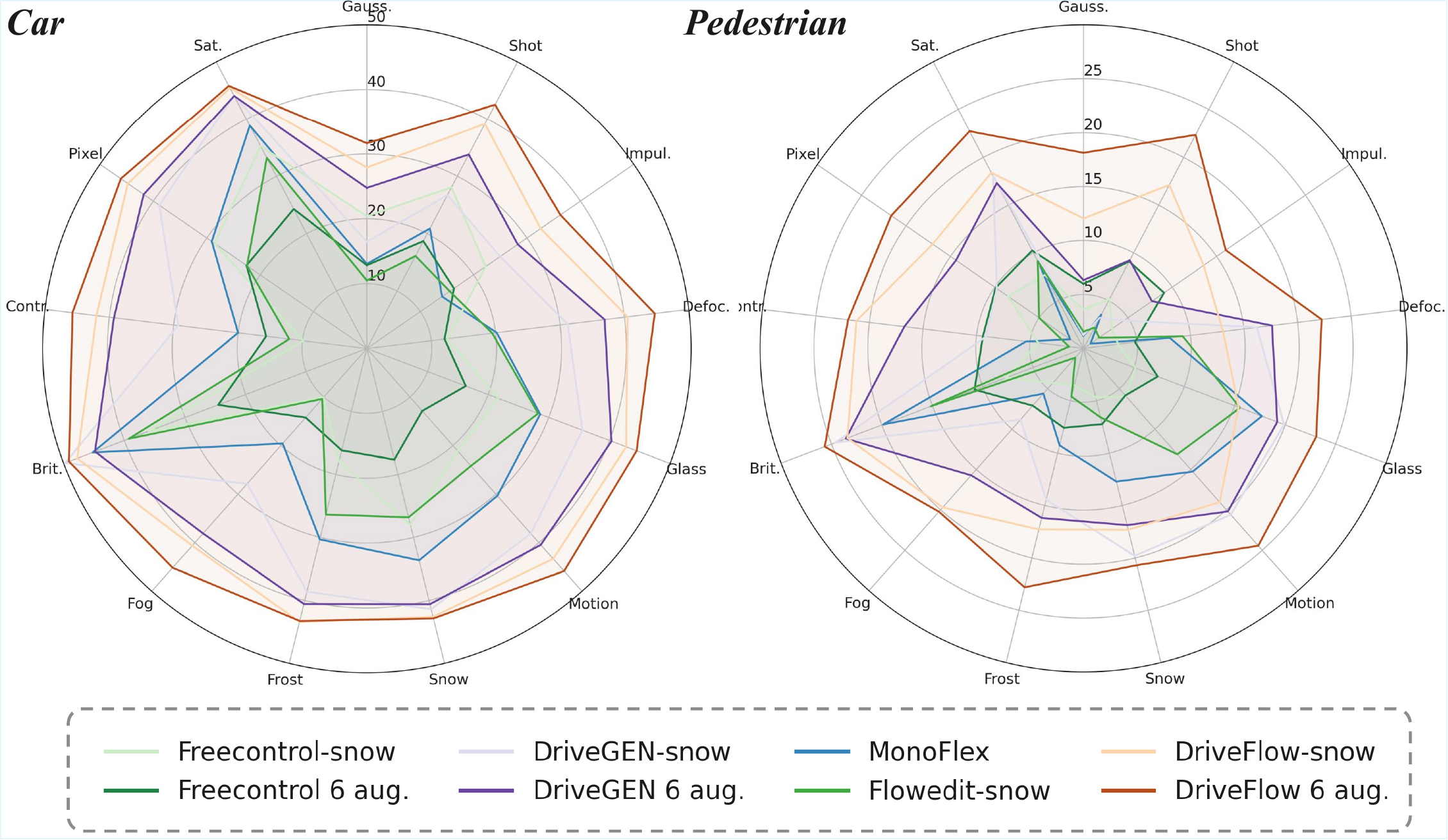}
\caption{Comparison on KITTI-C based on MonoFlex. \ournet~achieves 1) better performance with only Snow augmentation (orange) than DriveGEN with 6 aug. (purple) and
2) comprehensive gains on the minority class (Pedestrian) across OOD scenarios. Better viewed in color.
}
\label{fig:monoflex}
\end{figure}

\begin{figure*}[t]
\centering
\includegraphics[width=0.92\textwidth]{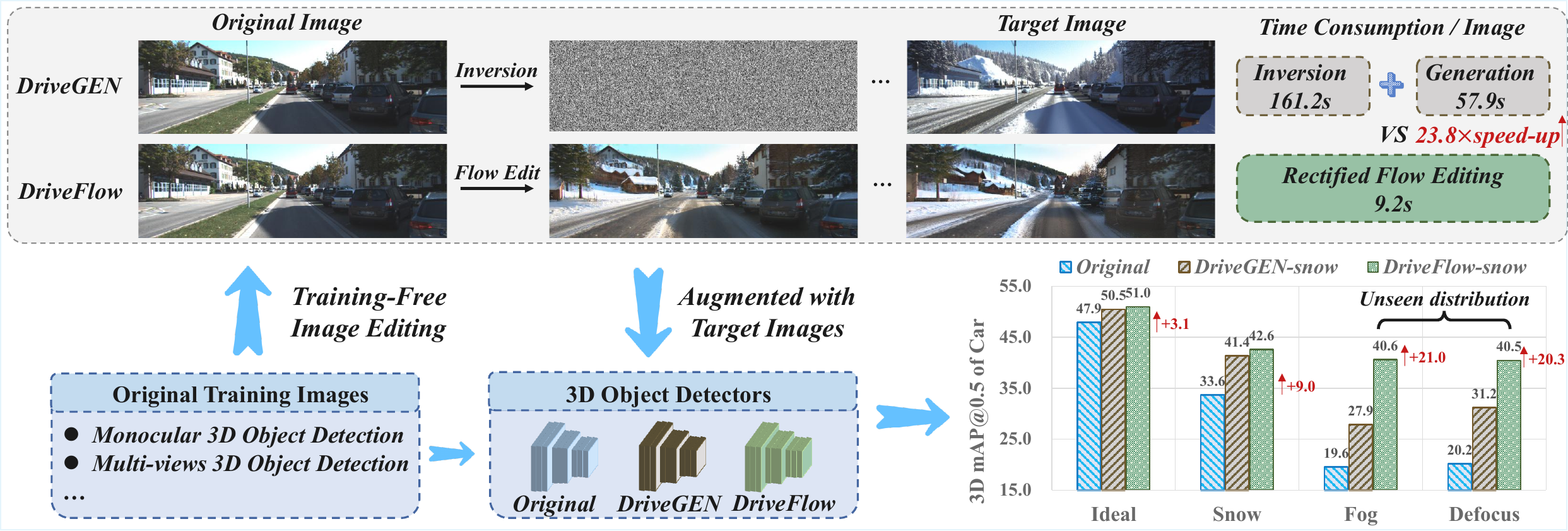}
\caption{An illustration of \ournet~for training data enhancement in vision-centric 3D object detection. 
In contrast to the inversion-based approach DriveGEN, \ournet~conducts rectified flow adaptation based on pre-trained T2I flow models (\eg Stable Diffusion 3), thereby 
achieving comprehensive improvement and rapid generation for 3D detectors.
}
\label{fig:teaser}
\end{figure*}

Such achievements mainly depend on one prerequisite: training data adequately covers all possible test scenarios.
However, it is particularly challenging to satisfy this assumption since driving systems often operate continuously outdoors over extended periods. 
Once the system suffers from unexpected data changes, well-trained detectors often fail to maintain the performance due to the shifts between training and test data distributions, which is known as the out-of-distribution (OOD) issue~\cite{wang2020tent}.
To illustrate this, we follow DriveGEN~\cite{lin2025drivegen} and visualize the performance degradation of a well-trained detector when deployed across different environmental conditions, as shown in Figure~\ref{fig:teaser}.
The results clearly demonstrate that the detector achieves satisfactory performance under ideal conditions (daytime scenarios) while exhibiting significant performance deterioration in \emph{unseen} scenes (e.g., fog). 
Therefore, it is essential to enhance the robustness of 3D Object Detection models in systems, as unexpected performance degradation in OOD scenarios may pose severe safety risks.

\begin{figure*}[t]
\centering
\includegraphics[width=0.95\textwidth]{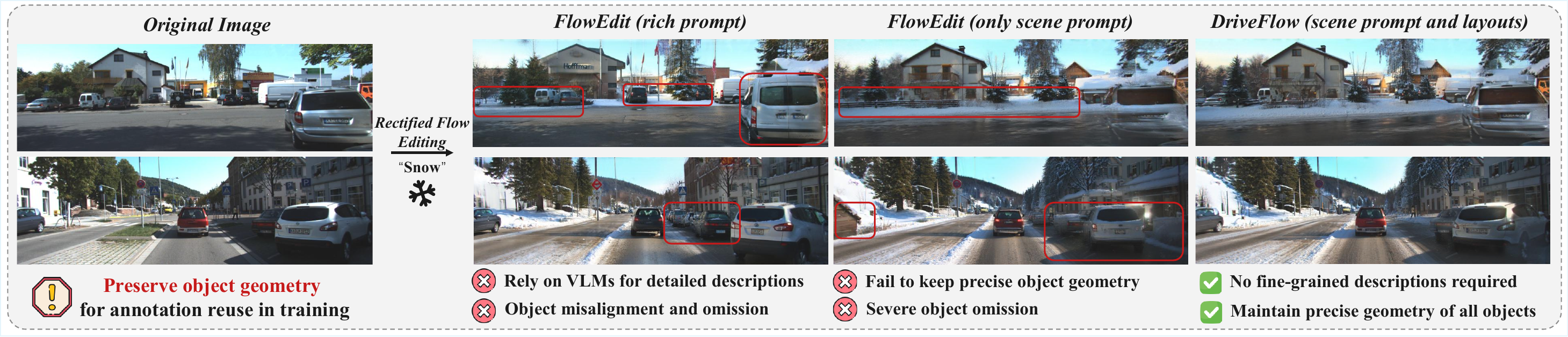}
\caption{
Due to the lack of foreground constraints, FlowEdit~\cite{kulikov2024flowedit} often fails to maintain 3D objects even with text descriptions from Qwen2.5-VL~\cite{bai2025qwen2}, while 
\ournet~only requires the target scene conditions and image layouts (\ie 2D bounding boxes).
Note that foreground preservation enables annotation reuse for augmented training.
}
\label{fig:motivation}
\end{figure*}

To handle the OOD issues in autonomous driving, previous approaches either rely on test-time model adaptation~\cite{lin2025monotta} or employ weather-adaptive diffusion models to transform adverse weather conditions to clear scenes~\cite{oh2025monowad}, which introduces additional computational cost at test time. 
Prior work DriveGEN~\cite{lin2025drivegen} employs controllable T2I diffusion generation to augment training data, thereby enhancing the robustness of 3D detectors. 
However, DriveGEN requires image inversion~\cite{song2020denoising} and relies on U-Net based pre-trained T2I diffusion models like Stable Diffusion 1.5~\cite{rombach2022high}. 
Previous methods~\cite{kulikov2024flowedit,wang2024taming} have shown that inversion-based editing produces unsatisfactory results regardless of whether ground-truth noise maps are available. 
Additionally, inversion-based approaches suffer from computational inefficiency (see  Figure~\ref{fig:teaser}) since reverting to noise maps requires more time compared to rectified-flow-based editing methods~\cite{kulikov2024flowedit}.
Recently, FlowEdit~\cite{kulikov2024flowedit} shows that leveraging pre-trained Text-to-Image (T2I) flow models (\eg Stable Diffusion 3~\cite{esser2024scaling} and FLUX~\cite{flux2024}) enables more powerful and efficient generation.
However, FlowEdit may pose potential risks of object misalignment and omissions even if fine-grained text descriptions are available, as shown in Figure~\ref{fig:motivation}.

To address these challenges, we propose an image editing method termed \emph{\textbf{\ournet}}, which is training-free and controllable based on pre-trained T2I flow models.
\ournet~aims to enhance training images in autonomous driving via performing frequency-based decomposition and adaptation of noise-free editing paths derived from velocities. 
Specifically, \ournet~consists of two strategies: 1) High-Frequency Foreground Preservation designs a foreground preservation loss for object regions to preserve accurate 3D geometry, while
2) Dual-Frequency Background Optimization introduces dual-frequency optimization to balance editing flexibility and semantic consistency of background regions.
As shown in Figure~\ref{fig:monoflex}, with only Snow augmentation, \ournet~performs better than six augmentations of DriveGEN, demonstrating more comprehensive robustness improvement across both the majority (\ie Car) and minority class (\ie Pedestrian).

\textbf{Contributions:} 
1) To the best of our knowledge, we are the first to apply rectified-flow-based editing for robust 3D object detection, offering novel perspectives on the usage of pre-trained T2I flow models in autonomous driving.
2) We propose \ournet~which incorporates high-frequency foreground preservation and dual-frequency background optimization strategies, achieving rapid (\eg 23.8x faster on KITTI) and effective (\eg 14.54 mAP improvement on KITTI-C with only snow augmentation) training data enhancement.
3) Extensive experiments validate that \ournet~brings comprehensive performance gains for both monocular and multi-view detectors. 
Moreover, \ournet~enhances robustness even for temporal-based 3D detectors, demonstrating our broad applicability.

\section{Related Work}
We first review model robustness studies for 3D detectors and controllable T2I diffusion methods. Additional discussions on vision-centric 3D detection are in Appendix \emph{\textbf{A}}.

\noindent
\textbf{Robust 3D Object Detection.}
Visual detection serves as a fundamental component in autonomous driving perception systems, enabling essential understanding of surroundings like traffic sign recognition.
Compared to LiDAR-based approaches, vision-centric 3D detectors offer lower hardware costs at the expense of model robustness, especially when encountering corrupted or out-of-distribution test data.
Recent approaches tackle this issue by: MonoWAD~\cite{oh2025monowad} adopts weather-adaptive diffusion models to revert weather conditions to ideal situations, whereas MonoTTA~\cite{lin2025monotta} improves model robustness via online test-time adaptation.
Additionally, MagicDrive~\cite{hong2021magicdrive}, Panacea~\cite{sun2022panacea}, and GAIA~\cite{hu2023gaia,russell2025gaia} leverage generative models to synthesize multi-view 3D driving scenes, addressing data scarcity in autonomous driving.
Despite their success, these methods introduce a considerable computational burden since they require substantial training data to train auxiliary modules or models.

\noindent
\textbf{Controllable T2I Image Diffusion.}
Pre-trained models such as Stable Diffusion~\cite{rombach2022high} and other large-scale architectures~\cite{ramesh2022hierarchical,flux2024} enable high-fidelity image synthesis.
This capability has improved controllable T2I diffusion to serve as a valuable paradigm for generating diverse synthetic data with fine-grained control. 
Recent methods such as ControlNet~\cite{zhao2023uni}, UniControl~\cite{qin2023unicontrol} and Layoutdiffusion~\cite{zheng2023layoutdiffusion} offer users spatial control based on trainable auxiliary modules. 
Alternatively, training-free methods like PnP~\cite{tumanyan2023plug} and FreeControl~\cite{mo2024freecontrol} manipulate self-attention features for semantic and spatial control. Besides, FlowEdit~\cite{kulikov2024flowedit} achieves the same goal in an inversion-free manner by constructing an ODE that directly maps source and target distributions.
However, even if fine-grained text descriptions are available (c.f. Figure~\ref{fig:motivation}), general-purpose editing methods still pose potential risks of object misalignment and omissions.
To solve it, DriveGEN~\cite{lin2025drivegen} extracts self-prototypes to guide the diffusion process for object preservation in autonomous driving.
Unfortunately, previous studies~\cite{kulikov2024flowedit,wang2024taming} have shown that inversion-based editing methods often suffer from unsatisfactory results and computational inefficiency.

\section{Preliminary}
\noindent
\textbf{Rectified Flow models.}
Flow-based generative models aim to construct a transportation between two distributions ${X}_0$ and ${X}_1$ through an ordinary differential equation (ODE): 
\begin{equation}
    dZ_t \;=\; V\!\left({Z}_t,t\right)\,dt,
\end{equation}
where time $t \in [0,1]$ and $V$ is a time-dependent velocity field which is typically parameterized by a learnable neural network.
The learned velocity field $V$ satisfies the boundary condition that if the vector ${Z}_1 \sim {X}_1$ at $t=1$, then ${Z}_0 \sim {X}_0$ at $t=0$.
Generally, we choose ${X}_1 = \mathcal{N}(0, {I})$ which allows to easily draw samples from the distribution ${X}_0$. 
To generate target samples, we get the initial Gaussian noise at $t=1$ and solve the ODE backward to $t=0$.

Rectified Flow~\cite{liu2022flow} is a particular paradigm of flow models, which learns a straight path to transport the Gaussian Noise distribution ${X}_1$ to the real data distribution ${X}_0$.
Thus, the marginal distribution ${X}_t$ at time $t$ corresponds to a linear interpolation between ${X}_0$ and ${X}_1$:
\begin{equation}
    {X}_t \sim (1-t){X}_0 + t{X}_1.
\end{equation}

With the text prompt $C$, T2I flow models adapt their velocity field $V$ to $V(X_t, t,C)$. Then, such models are trained on the image-text paired data $(X_0, C)$, which allows models to generate images via conditional sampling from $X_0| C$.

\begin{figure*}[t]
\centering
\includegraphics[width=0.95\textwidth]{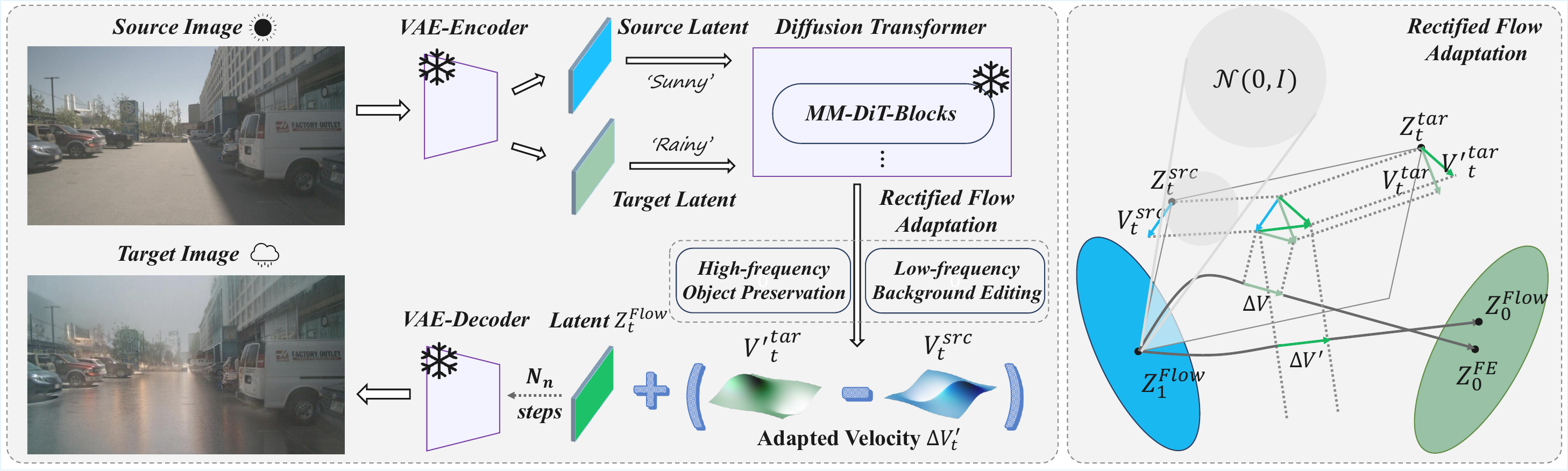}
\caption{An illustration of \ournet.
Without modification of the pre-trained model, \ournet~employs frequency-based decomposition for both velocity fields $V_t^{src}$ and $V_t^{tar}$, and then applies: 
1) High-Frequency Foreground Preservation, applying a L2 alignment loss to align high-frequency contents between velocity fields explicitly.
2) Dual-Frequency Background Optimization, introducing dual-frequency optimization for background areas to ensure editing flexibility and semantic consistency.
}
\label{fig:method}
\end{figure*}

\noindent
\textbf{FlowEdit.}
Inversion-based editing involves two stages: 1) Invert the source image to noise space via the forward trajectory $Z_t^{src}$, then 2) generate the target image from the noise latent via the reverse trajectory $Z_t^{tar}$.
FlowEdit~\cite{kulikov2024flowedit} shows this process can be reformulated as a direct path $Z_t^{{inv}} = Z_0^{{src}} + Z_t^{{tar}} - Z_t^{{src}}$.
This equation can be further expressed as an ODE:  
\begin{equation}
    dZ_t^{{inv}} = V_t^{\triangle}(Z_t^{{src}}, Z_t^{{inv}} + Z_t^{{src}} - Z_0^{{src}}) dt.
\label{eqn:init_inv}
\end{equation}

Since a fixed $Z_t^{{src}}$ often create mismatched pairings, FlowEdit solves it by averaging across multiple random pairings: $\hat{Z}_t^{\mathrm{src}} = (1-t)Z_0^{\mathrm{src}} + tN_t$ where $N_t \sim \mathcal{N}(0, 1)$.
Therefore, substituting back into the Eqn.~(\ref{eqn:init_inv}), we obtain:
\begin{equation}
    dZ_t^{\mathrm{FE}} = \mathbb{E}\left[V_t^{\Delta}(\hat{Z}_t^{\mathrm{src}}, Z_t^{\mathrm{FE}} + \hat{Z}_t^{\mathrm{src}} - Z_0^{\mathrm{src}})\,\middle|\,Z_0^{\mathrm{src}}\right] dt.
\end{equation}
This path achieves \emph{noise-free} editing since the velocity difference vector $V_t^{\triangle}(\hat{Z}_t^{{src}}, \hat{Z}_t^{{tar}})$ cancels out the same noise (c.f. the light green arrow in Figure~\ref{fig:method}). 
In this way, noise-free trajectories enhance editing stability by preventing stochastic disturbances during the generation process.

\section{Rectified Flow Adaptation}
\noindent
\textbf{Problem Statement.}
Based on the labeled training images $\mathcal{D}{=}\{(\mathbf{x}_i, \mathbf{y}_i)\}_{i=1}^{N}$, we can obtain the well-trained 3D visual detector $f_{\Theta_{d}}(\cdot)$ where $\Theta_{d}$ represents the learnable parameters.
A total of $N$ training images are drawn from the training distribution $P\left(\mathbf{x}\right)$ (\ie,~$\mathbf{x} \sim P\left(\mathbf{x}\right)$).
During deployment, the model accesses unlabeled test images $\mathcal{D}_t{=}\{\mathbf{x}^t_i\}_{i=1}^{M}$ from distribution $Q(\mathbf{x})$ (\ie~$\mathbf{x}^t \sim Q\left(\mathbf{x}\right)$), which often differs from the training distribution $P(\mathbf{x})$ due to diverse environmental conditions and weather variations, \ie~$P\left(\mathbf{x}\right) \neq Q\left(\mathbf{x}\right)$. 
Once data distribution shifts exist, the well-trained detector encounters the  Out-of-distribution (OOD) issue, leading to unexpected performance drop.

Prior works suffer from several key challenges in addressing the OOD issue. 
General image editing methods~\cite{mo2024freecontrol,kulikov2024flowedit} fail to maintain all objects with precise geometry, while test-time approaches
~\cite{lin2025monotta,oh2025monowad} require additional computation costs during inference.
Prior method~\cite{lin2025drivegen} relies on inversion-based techniques, which may lead to suboptimal results and a significant computational burden.

\noindent
\textbf{Overall Scheme.}
We introduce \ournet, a rectified flow adaptation method for training data enhancement in vision-centric 3D Object Detection, which builds upon pre-trained T2I flow models as illustrated in Figure~\ref{fig:method}. The editing process is driven by a set of timesteps $\{ t_i \}_{i=0}^{T}$ where $T$ represents the total number of  intervals.
The objective of \ournet~is to learn a suitable target velocity ${V^{\prime}}_t^{tar}$ through $N_n$ inner iterations at each of the $N_{max}$ diffusion steps, subject to the constraint $N_{max} \leq T$.

Without loss of generality, given a source image $X_0^{src}$, we first encode it by the encoder of the Variational AutoEncoder (VAE) to obtain the initial latent $Z_0^{src}$.
Then, we prepare two latent-prompt pairs for the diffusion transformer.
For the source pair, the source latent at time $t_i$ is equal to:
\begin{align}
\label{eqn:srcz}
    \hat{Z}_{t_i}^{{src}} = (1-t_i)Z_0^{{src}} + t_iN_t.
\end{align}
The source text prompt ${c}_{src}$ is generated by simply describing the scene of $X_0^{src}$ (\eg `An urban scene on a sunny day').
As for the target pair, the target latent is obtained by:
\begin{align}
\label{eqn:tarz}
    \hat{Z}_{t_i}^{{tar}} = Z_{t_i}^{{Flow}} + \hat{Z}_{t_i}^{{src}} - Z_0^{{src}},
\end{align}
where $Z_{t_{max}}^{{Flow}}$ is initialized by $Z_0^{src}$ and $t_{{max}} = \max_i \{ t_i \}$.
Similarly, the target prompt ${c}_{tar}$ contains the description for the desired scene (\eg rainy).
Given the source and target velocity fields $V^{src}_t$ and $V^{tar}_t$, \ournet~performs frequency-based decomposition, parameterizing $V^{tar}_t$ as a learnable vector.
To learn an appropriate ${V^{\prime}}_t^{tar}$, the key challenge lies in achieving the desired scene-level editing without compromising the integrity of the 3D object geometry.
To this end, \ournet~first applies the foreground preservation loss $\mathcal{L}_{{obj}}$ between the high‑frequency components of the foreground to maintain 3D object geometry.
Meanwhile, \ournet~derives a spatial cosine‑similarity map between the low‑frequency components for background regions, utilizing it to compute the diversity loss  $\mathcal{L}_{{div}}$ for sufficient editing intensity.
To prevent unexpected collapse on background regions, \ournet~also enables the high-frequency background regularization term $\mathcal{L}_{{bg}}$ (see Algorithm~\ref{al:adapt}).

Overall, the total scheme of \ournet~is as follows:
\begin{align}
\label{eqn:total_loss}
    \mathcal{L}_{total} = \lambda_1\mathcal{L}_{{obj}} + \lambda_2\mathcal{L}_{{div}} + \lambda_3\mathcal{L}_{{bg}},
\end{align}
where $\lambda_1,\lambda_2,\lambda_3$ are hyper-parameters.
Subsequently, we obtain the updated target velocity ${V^{\prime}}_t^{tar}$ which guides the velocity difference $\Delta {V_t'}$ via:
\begin{align}
\label{eqn:delta_v}
    V_t^{\triangle} \leftarrow {V^{\prime}}_t^{tar}(\hat{Z}^{{tar}}_{t_i}, t_i)-V^{src}(\hat{Z}^{{src}}_{t_i}, t_i).
\end{align}
Eventually, we update the edited latent ${Z}^{{Flow}}_{t_{i-1}}$ via:
\begin{align}
\label{eqn:update_z}
    {Z}^{{Flow}}_{t_{i-1}} \leftarrow {Z}^{{Flow}}_{t_i} + (t_{i-1} - t_i){V}^{\Delta}_{t_i}.
\end{align}

\subsection{High-Frequency Foreground Preservation} 
General-purpose editing methods~\cite{kulikov2024flowedit,mo2024freecontrol}
 often fail to maintain 3D object geometry (see figure~\ref{fig:motivation}) even when guided by detailed text descriptions from Qwen2.5-VL~\cite{bai2025qwen2}.
To handle this, we decompose the velocity fields $V^{src}_t$ and $V^{tar}_t$ at timestep $t$ by applying
the Gaussian blur $G_\sigma^{(k)}$ to achieve the low-frequency components $V_{L,t}$:
\begin{align}
\label{eqn:lf}
    % \text{low}(V;\sigma) &= V * G_\sigma^{(k)}, \\
    V_{L,t} &= V * G_\sigma^{(k)}, \\
    G_\sigma^{(k)}(i,j) &= \frac{1}{K} \exp\left(-\frac{i^2 + j^2}{2\sigma^2}\right),
\end{align}
where $i,j \in \left\{-\frac{k-1}{2}, ..., \frac{k-1}{2}\right\}$ depends on the kernel size $k$, $K$ is a constant and $\sigma$ controls the blur strength.
Therefore, we can achieve the high-frequency component $V_{H,t}$ by:
\begin{align}
\label{eqn:hf}
    V_{H,t} = V - V_{L,t}.
\end{align}
Since $G_\sigma^{(k)}$ acts as a low-pass filter to preserve slowly varying components $V_{L,t}$, the high-frequency residual $V_{H,t}$ in Eqn.~\ref{eqn:hf} captures rapidly varying components that typically correspond to objects within 2D bounding boxes.
With parameterization of $V^{tar}$ as a learnable vector, we calculate the foreground preservation loss $\mathcal{L}_{{obj}}$ between $V_{H,t}^{src}$ and $V_{H,t}^{tar}$ within all object regions: 
\begin{align}
\label{eqn:loss_obj}
    \mathcal{L}_{{obj}} = \frac{1}{|\mathbf{M}|} \|\mathbf{M} \odot (V_{H,t}^{{tar}} - V_{H,t}^{{src}})\|_2^2,
\end{align}
where $\mathbf{M}$ is the binary mask derived from the coordinate transformation of image layouts $\mathbf{L}$ through downsampling, with object regions marked as 1 and background as 0.

\subsection{Dual-Frequency Background Optimization}
To fully exploit the pre-trained T2I flow model, we aim for sufficient editing intensity in background regions.
To this end, we first compute the diversity loss $\mathcal{L}_{{div}}$ between the low-frequency components $V_{L,t}^{{src}}$ and $V_{L,t}^{{tar}}$ within background regions by:
\begin{align}
\label{eqn:loss_div}
\mathcal{L}_{{div}} = \frac{1}{|\bar{\mathbf{M}}|} \sum_{\bar{\mathbf{M}}} {\cos}(V_{L,t}^{{tar}}, V_{L,t}^{{src}}),
\end{align}
where $\bar{\mathbf{M}}=1-{\mathbf{M}}$ and $\cos(\mathbf{a}, \mathbf{b}) = \frac{\mathbf{a} \cdot \mathbf{b}}{\|\mathbf{a}\|_2 \; \|\mathbf{b}\|_2} \in [-1, 1]$ denotes the cosine-similarity calculation.
Specifically, the objective of the diversity loss $\mathcal{L}_{{div}}$ is to maximize the discrepancy between the source and target low-frequency components of background regions, which encourages the optimized velocity field ${V^{\prime}}_t^{tar}$ to exhibit sufficient variations.
By emphasizing regions with higher similarity, $\mathcal{L}_{{div}}$ guides this process to pay more attention to the regions which are more similar to the original ones. Such a design encourages more comprehensive background editing.

However, exclusive reliance on $\mathcal{L}_{div}$ for the velocity filed adaptation may result in trivial solutions within the background, \ie the optimized velocity field indiscriminately seeks to maximize the differences from the source velocity field $V_{L,t}^{{src}}$.
To prevent the unexpected collapse, we further enforce semantic consistency constraints by applying the background regularization term:
\begin{align}
\label{eqn:loss_bg}
    \mathcal{L}_{{bg}} = \frac{1}{|\bar{\mathbf{M}}|} \|\bar{\mathbf{M}} \odot (V_{H,t}^{{tar}} - V_{H,t}^{{src}})\|_2^2.
\end{align}
With the introduction of $\mathcal{L}_{{bg}}$, the background editing process achieves a trade-off between diversity and semantic consistency. This dual-frequency background optimization mechanism ensures the simultaneous achievement of background diversity and semantic consistency, thereby effectively mitigating potential semantic drift. Prior approach~\cite{lin2025drivegen} often emphasizes explicit foreground constraints while overlooking the need for semantic consistency in background regions during the editing process.
However, it is essential for temporal multi-view 3D object detection~\cite{huang2022bevdet4d} to apply reasonable constraints to background regions since it controls whether the augmented training data retains adequate temporal consistency (c.f. Temporal-Based Section in Experiments). We summarize 
the Pseudo-code of \ournet~in Algorithm~\ref{al:adapt}.

\begin{algorithm}[t]
% \scriptsize
    \caption{The pipeline of the proposed \ournet}\label{al:adapt}
    \begin{algorithmic}[1]
        \REQUIRE Training data $\{(\mathbf{x}_i^s, \textbf{y}_i^s)\}_{i=1}^{N}$; Hyper-parameters $\lambda_1,\lambda_2,\lambda_3,N_n,N_{max}$; Target scene; Pre-trained model.
    \FOR {each training image $\mathbf{x}_i$} 
        \FOR {diffusion step $i = N_{max} \to 1$}
            \STATE Get $c_{src}$, $c_{tar}$ based on the source and target scene;
            \STATE Extract the source latent $Z_0^{src}$ and initialize $Z_{t_i}^{{Flow}}$;
            \STATE Get $\hat{Z}_{t_i}^{\mathrm{src}}$ and $\hat{Z}_{t_i}^{{tar}}$ based on Eqn.~\ref{eqn:srcz} and Eqn.~\ref{eqn:tarz};
            \STATE Undergo the transformer to get $V^{src}$ and $V^{tar}$;
            \STATE Decomposition based on Eqn.~\ref{eqn:lf} and Eqn.~\ref{eqn:hf};
            \FOR {inner loop $n = 1 \to N_n$}
                \STATE Calculate the loss terms $\mathcal{L}_{{obj}},\mathcal{L}_{div},\mathcal{L}_{bg}$ based on Eqn.~\ref{eqn:loss_obj}, Eqn.~\ref{eqn:loss_div} and Eqn.~\ref{eqn:loss_bg};
                \STATE Update ${V^{\prime}}_t^{tar}$ based on Eqn.~\ref{eqn:total_loss};
            \ENDFOR
            \STATE Update $V_t^{\triangle}$ based on Eqn.~\ref{eqn:delta_v};
            \STATE Update ${Z}^{{Flow}}_{t_{i-1}}$ based on Eqn.~\ref{eqn:update_z};
        \ENDFOR
    \ENDFOR
    \RETURN Output images for all $\mathbf{x}_i$ of the target scene.
     \end{algorithmic}
\end{algorithm}

\begin{table*}[t]
    \begin{center}
    \fontsize{9pt}{\baselineskip}\selectfont
    \setlength\tabcolsep{3pt}

    \begin{tabular}{l|ccc|ccc|cccc|ccc|c}
         \toprule
         \multicolumn{15}{c}{{{Car}, IoU @ 0.7, 0.5, 0.5}} \\
         \midrule
         \multirow{2}{*}{Method}  &
         % \multirow{2}{*}{\shortstack{Inversion-\\free}} &
         % \multirow{2}{*}{\shortstack{Training-\\free}} &
         \multicolumn{3}{c|}{Noise} & 
         \multicolumn{3}{c|}{Blur} & 
         \multicolumn{4}{c|}{Weather} & 
         \multicolumn{3}{c|}{Digital} & 
         \multirow{2}{*}{Avg.} \\
        \cmidrule(lr){2-4} \cmidrule(lr){5-7} \cmidrule(lr){8-11} \cmidrule(lr){12-14} 
        & Gauss. & Shot & Impul. & Defoc. & Glass & Motion & Snow & Frost & Fog & Brit. & Contr. & Pixel & Sat. \\
         \midrule
        MonoGround %~\cite{qin2022monoground}
        & 13.05 & 21.77 & 18.87 & 20.79 & 30.74 & 32.02 & 34.43 & 27.02 & 14.15 & 46.21 & 14.63 & 33.41 & 35.60 & 26.36  \\
        \midrule
        
        ~$\bullet~$ Color Jitter (\textit{Trad.}) & 12.88 & 24.31 & 18.95 & 23.07 & 30.44 & 31.42 & 35.94 & 30.43 & 19.89 & 44.66 & 20.61 & 29.75 & 36.65 & 26.36 \\
        ~$\bullet~$ Brightness (\textit{Trad.}) & 14.02 & 23.52 & 20.14 & 23.95 & 31.78 & 28.79 & 35.08 & 31.87 & 18.87 & 42.94 & 17.75 & 25.55 & 37.18 & 27.03 \\
        
        \midrule
        ~$\bullet~$ ControlNet (\textit{Snow})  &  1.76 & 3.23 & 4.63 & 5.20 & 12.95 & 14.11 & 17.70 & 11.58 & 3.04 & 35.21 & 2.98 & 7.29 & 13.98 & 10.28  \\  
        ~$\bullet~$ ControlNet (\textit{6 × Aug.})  & 0.00 & 0.00 & 0.00 & 1.68 & 1.26 & 0.35 & 1.13 & 0.52 & 0.44 & 4.08 & 0.38 & 2.22 & 1.77 & 1.06  \\  
         \midrule
        ~$\bullet~$ FreeControl (\textit{Snow})  & 11.75 & 21.89 & 15.76 & 17.70 & 21.45 & 21.69 & 32.08 & 20.60 & 13.57 & 36.05 & 14.03 & 26.75 & 38.35 & 22.43  \\  
        ~$\bullet~$ FreeControl (\textit{6 × Aug.}) & 15.20 & 22.59 & 15.35 & 22.00 & 21.18 & 18.95 & 17.69 & 14.85 & 14.82 & 24.02 & 16.97 & 22.99 & 26.12 & 19.44 \\  
         % \midrule
        ~$\bullet~$ DriveGEN (\textit{Snow})  & 17.07 & 26.78 & 23.78 & 32.89 & 37.52 & 39.06 & {40.61} & 34.91 & 25.29 & {46.21} & 27.12 & 38.25 & 44.45 & 33.38 \\  
        ~$\bullet~$ DriveGEN (\textit{6 × Aug.})  & {23.84} & {32.59} & {30.34} & {38.57} & {41.20} & {40.19} & 38.16 & {38.40} & {32.53} & 43.95 & {34.80} & {44.10} & {45.13} & {37.21}  \\ 
        \midrule
        ~$\bullet~$ FlowEdit (\textit{Snow}) & 4.38 & 8.54 & 6.98 & 24.57 & 30.98 & 27.19 & 27.84 & 28.36 & 24.32 & 38.31 & 28.84 & 28.00 & 31.98 & 23.87  \\ 
        ~$\bullet~$ \ournet~(\textit{Snow}) & 26.73 & 35.70 & 26.59 & 38.22 & 41.73 & 42.16 & \textbf{43.43} & 40.73 & 41.20 & 47.16 & 43.72 & 44.15 & 45.18 & 39.75  \\ 
        ~$\bullet~$ \ournet~(\textit{6 × Aug.}) & \textbf{29.64} & \textbf{39.45} & \textbf{30.56} & \textbf{43.95} & \textbf{45.02} & \textbf{45.49} & 42.63 & \textbf{42.51} & \textbf{44.18} & \textbf{47.73} & \textbf{45.61} & \textbf{46.59} & \textbf{46.22} & \textbf{42.27}  \\

        \midrule
        \midrule

        MonoCD %~\cite{qin2022monoground}
        & 8.88 & 15.60 & 13.22 & 23.44 & 32.83 & 33.93 & 30.18 & 27.94 & 22.52 & 46.07 & 23.20 & 29.87 & 37.31 & 26.54   \\
        \midrule
        ~$\bullet~$ Color Jitter (\textit{Trad.}) & 8.61 & 14.28 & 12.79 & 21.13 & 32.22 & 33.81 & 32.14 & 30.63 & 24.03 & 45.09 & 25.68 & 30.57 & 38.78 & 26.90\\
        ~$\bullet~$ Brightness (\textit{Trad.}) & 11.76 & 19.38 & 16.09 & 21.60 & 31.01 & 32.36 & 32.32 & 29.87 & 22.56 & 45.69 & 24.56 & 34.70 & 39.18 & 27.78 \\
        \midrule
        ~$\bullet~$ ControlNet (\textit{Snow})  & 0.00 & 0.00 & 0.00 & 1.00 & 1.59 & 4.35 & 5.06 & 5.99 & 2.67 & 18.24 & 3.28 & 0.64 & 1.57 & 3.41 \\
        ~$\bullet~$ ControlNet (\textit{6 × Aug.})  & 0.00 & 0.00 & 0.00 & 0.00 & 0.00 & 0.00 & 0.00 & 0.00 & 0.00 & 0.00 & 0.00 & 0.00 & 0.00 & 0.00  \\
        \midrule
        ~$\bullet~$ FreeControl (\textit{Snow})  & 11.30 & 20.10 & 13.00 & 16.10 & 23.70 & 24.20 & 27.70 & 22.60 & 19.60 & 32.20 & 20.90 & 30.00 & 34.50 & 22.80\\
        ~$\bullet~$ FreeControl (\textit{6 × Aug.})  & 12.90 & 20.00 & 13.00 & 13.60 & 16.70 & 14.60 & 15.70 & 13.50 & 15.60 & 21.30 & 15.60 & 21.60 & 23.10 & 16.70 \\
        ~$\bullet~$ DriveGEN (\textit{Snow})  & 19.91 & 28.93 & 24.87 & 35.06 & 38.61 & 38.81 & 37.00 & 37.32 & 37.26 & 43.74 & 38.37 & 41.86 & 43.56 & 35.79 \\
        ~$\bullet~$ DriveGEN (\textit{6 × Aug.})  &  23.35 & 34.49 & 30.36 & 40.47 & 41.15 & 42.67 & 40.08 & 39.61 & 41.51 & 46.15 & 42.99 & 44.35 & 45.57 & 39.44  \\ 
        \midrule
        ~$\bullet~$ FlowEdit (\textit{Snow}) & 8.43 & 14.94 & 7.14 & 28.44 & 33.26 & 32.53 & 30.95 & 30.63 & 35.66 & 40.40 & 36.82 & 32.82 & 39.26 & 28.56 \\
        ~$\bullet~$ \ournet~(\textit{Snow}) & 27.84 & 39.42 & 30.02 & 38.92 & 42.18 & 43.82 & \textbf{41.64} & \textbf{42.43} & 43.36 & \textbf{47.53} & 44.16 & 45.22 & \textbf{46.38} & 40.99 \\
        ~$\bullet~$ \ournet~(\textit{6 × Aug.}) &  \textbf{29.26} & \textbf{40.64} & \textbf{32.04} & \textbf{44.55} & \textbf{44.90} & \textbf{44.81} & 39.76 & 39.53 & \textbf{45.50} & 47.30 & \textbf{46.00} & \textbf{45.83} & 45.84 & \textbf{42.00} \\

         \bottomrule
         \end{tabular}

    \end{center}
        \caption{\label{tab:kitti-c} Comparison on {KITTI-C}, severity {level 1} regarding {Mean $AP_{3D|R_{40}}$}. The \textbf{bold} number indicates the best result.
    }
\end{table*}

\section{Experiments}
We validate the effectiveness for \ournet~on both monocular and multi-view 3D object detection.
Following DriveGEN~\cite{lin2025drivegen}, we set three different training settings with various enhanced scenarios:
1) Traditional techniques (\ie Color Jitter and Brightness);
2) Scenes with Snow augmentation;
% 3) Scenes with Snow, Rain and Fog augmentation (\textit{3 × Aug.});
3) Scenes with Snow, Rain, Fog, Night, Defocus and Sandstorm augmentation (\textit{6 × Aug.}).
More implementation details are put in Appendix \emph{\textbf{B}}.

\noindent
\textbf{Datasets.}
In monocular 3D object detection, we follow the existing protocol~\cite{zhang2021objects} to split the images of KITTI~\cite{geiger2012we} into a training set (3712 images) and a validation set (3769 images), including three classes: Car, Pedestrian, and Cyclist.
To validate the model robustness, well-trained detectors are evaluated on KITTI-C~\cite{lin2025monotta}, including 13 corrupted scenarios for validation across four categories: Noise, Blur, Weather, and Digital~\cite{hendrycks2018benchmarking}.

For multi-view 3D object detection, we conduct experiments on the nuScenes~\cite{caesar2020nuscenes} dataset. 
Following DriveGEN~\cite{lin2025drivegen}, we augment 500 daytime training scenes under the \emph{snow} condition to enhance multi-view 3D detectors. Then, they are evaluated on the widely used Robo3D benchmark~\cite{xie2025benchmarking}. 
Moreover, we also validate \ournet~for enhancing temporal-based methods on real-world scenarios following~\cite{liu2023bevfusion}. More dataset details are provided in Appendix~\emph{\textbf{C}}.

\noindent\textbf{Compared Methods.}
All the experiments are based on well-known or state-of-the-art baselines~\cite{zhang2021objects,qin2022monoground,yan2024monocd,li2022bevformer,huang2022bevdet4d}.
We compare \ournet~with:
1) Well-trained model, \ie fully trained on original data and apply the model to corrupted test data;
2) Traditional data augmentation techniques, \ie Color Jitter and Brightness;
3) Training-based T2I diffusion: ControlNet~\cite{zhang2023adding} with additional masks~\cite{ravi2024sam} and prompts~\cite{chen2024internvl};
4) Training-free T2I diffusion (inversion-based): FreeControl~\cite{mo2024freecontrol} and DriveGEN~\cite{lin2025drivegen} enables zero-shot control of pretrained diffusion models.
5) Rectified-flow editing (inversion-free): FlowEdit~\cite{kulikov2024flowedit} enables powerful generation based on pre-trained T2I flow models.

\noindent\textbf{Evaluation Protocols}.
For monocular 3D object detection, we primarily report experimental results using Average Precision (AP) for 3D bounding boxes, denoted as $AP_{3D|R_{40}}$. On the KITTI-C dataset, results on the KITTI-C dataset are averaged across three difficulty levels, with Intersection over Union (IoU) thresholds set to 0.7, 0.5, 0.5 for Cars and 0.5, 0.25, 0.25 for Pedestrians and Cyclists, respectively.
As for multi-view 3D object detection, we report the mean average precision (mAP) and nuScenes detection score (NDS).

% \noindent\textbf{Implementation Details.}
% We implement all methods in PyTorch~\cite{paszke2019pytorch}. Following FlowEdit~\cite{kulikov2024flowedit}, we adopt Stable-Diffusion-3-medium~\cite{esser2024scaling} and set $N_{max}=33$ and $T=50$.
% We set $\lambda_1=5,\lambda_2=1,\lambda_3=1,N_n=5$.
% All 3D detectors are trained on a simple combination of original and augmented data.

\subsection{Comparisons with Previous Methods}
In monocular 3D object detection, the results of Figure~\ref{fig:monoflex} and Table~\ref{tab:kitti-c} reveal that:
1) Well-trained detectors exhibit substantial performance degradation when deployed under corrupted scenarios, and conventional augmentation techniques fail to mitigate the data distribution shifts.
2) Due to the absence of foreground constraints, ControlNet~\cite{zhang2023adding}, FreeControl~\cite{mo2024freecontrol} and FlowEdit~\cite{kulikov2024flowedit} yield marginal improvements for the single snow augmentation.
As more augmented scenes are incorporated, they show progressively declining performance.
3) \ournet~consistently outperforms DriveGEN~\cite{lin2025drivegen} within all baselines across 13 OOD scenarios. Remarkably, with only a single snow augmentation, our method outperforms DriveGEN with six augmented scenes across both majority and minority classes.

For multi-view 3D object detection, we follow the DriveGEN protocol by selecting 3,000 daytime training images and applying the snow augmentation (3k Snow) for enhancement.
Table~\ref{tab:nus} shows that \ournet~enhances BEVFormer-tiny~\cite{li2022bevformer} to achieve better performance, outperforming DriveGEN across all 8 OOD scenarios in nuScene-C~\cite{xie2025benchmarking}. 
Considering the substantial computational efficiency of \ournet, these results further validate our effectiveness. More results are put in Appendix \emph{\textbf{D}}.

\begin{figure*}[t]
\centering
\includegraphics[width=0.8\textwidth]{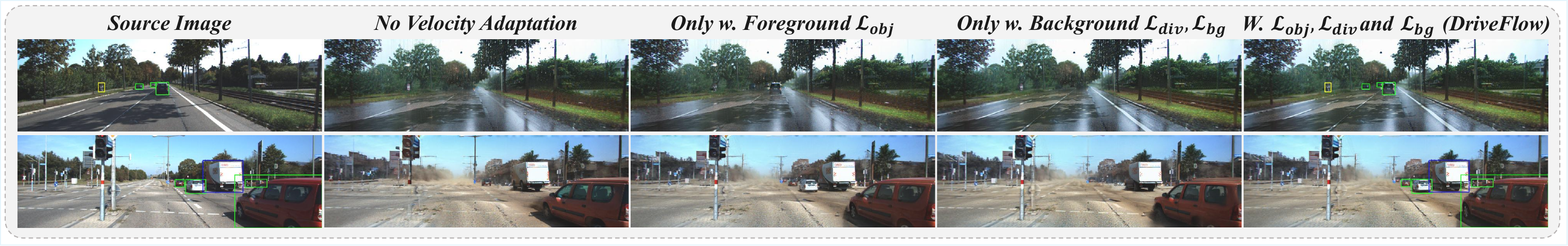}
\caption{Ablation studies on the loss terms  $\mathcal{L}_{{obj}}$,  $\mathcal{L}_{{div}}$ and $\mathcal{L}_{{bg}}$. More results are available in Appendix \emph{\textbf{E}}.} 
\label{fig:abla}
\end{figure*}

\begin{figure*}[t]
\centering
\includegraphics[width=0.8\textwidth]{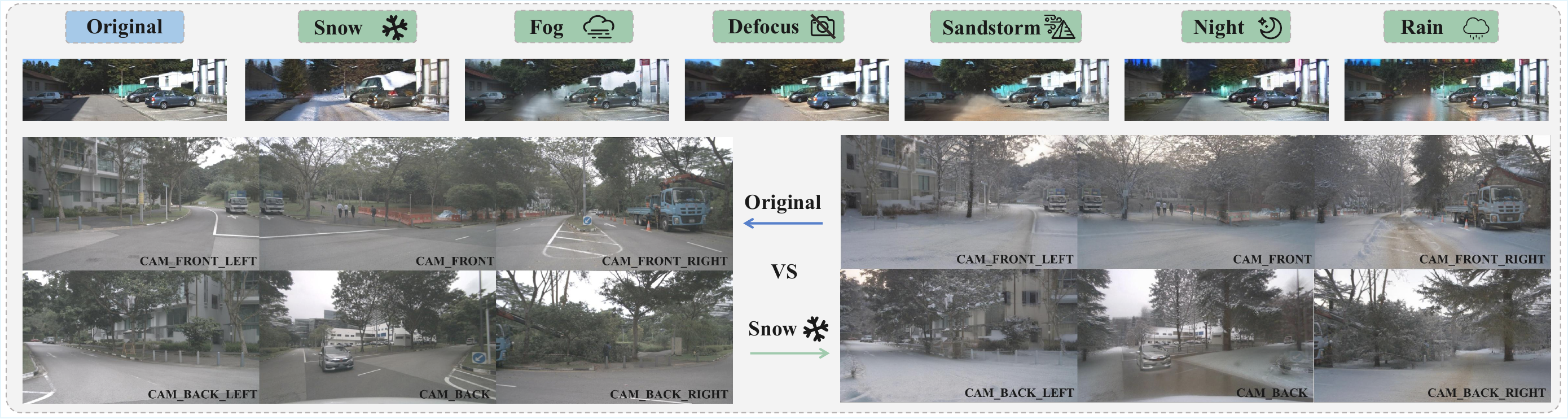}
\caption{Qualitative visualizations of \ournet~based on KITTI with various scenes and nuScenes with different views.}
\label{fig:vis}
\end{figure*}

\begin{table*}[!h]
  \setlength\tabcolsep{3pt}
  \fontsize{9pt}{\baselineskip}\selectfont
  \centering

  \begin{tabular}{l|l|cccccccccccc}
    \toprule
    \multirow{2}{*}{Metric} & \multirow{2}{*}{Method} & \multicolumn{9}{c}{nuScenes-C} \\
    \cmidrule(lr){3-11} 
    & & Brightness & CameraCrash & ColorQuant & Fog & FrameLost & LowLight & MotionBlur & Snow & Avg. \\
    \midrule
    \multirow{3}{*}{mAP} & BEVFormer-tiny & 24.26 & 14.89 & 23.91 & 21.98 & 20.43 & 16.11 & 20.78 & 10.83 & 19.15 \\
    & ~$\bullet$ DriveGEN (\textit{3k Snow}) & 26.04  & 15.79  & 25.78  & 24.13  & 21.13  & 17.20  & 22.39  & 11.72  & 20.52  \\
    % & ~$\bullet~$ FlowEdit (\textit{3k Snow}) \\
    & ~$\bullet~$\ournet~(\textit{3k Snow}) & \textbf{26.26} & \textbf{16.76} & \textbf{25.82} & \textbf{24.45} & \textbf{21.72} & \textbf{17.87} & \textbf{22.53 }& \textbf{12.62} & \textbf{21.00} \\
    % \midrule
    % \multirow{3}{*}{NDS} & BEVFormer-tiny & 34.93 & 28.04 & 34.56 & 33.04 & 31.42 & 28.42 & 31.81 & 23.27 & 30.69 \\
    % & ~$\bullet$ DriveGEN (\textit{3k Snow}) & 37.14  & 30.02  & \textbf{36.93}  & 35.35  & \textbf{33.32}  & 30.42  & 34.23  & 24.09  & 32.69  \\
    % % & ~$\bullet~$ FlowEdit (\textit{3k Snow}) \\
    % & ~$\bullet~$\ournet~(\textit{3k Snow}) & \textbf{37.20}  & \textbf{30.64} & 36.52 & \textbf{35.93} & 33.09 & \textbf{30.71} & \textbf{34.28} & \textbf{25.05} & \textbf{32.93}  \\  
    \bottomrule
  \end{tabular}
    \caption{\label{tab:nus} Detection results on nuScenes-C, regarding mAP. Due to page limitations, results in terms of NDS are in Appendix \emph{\textbf{D}}.}
\end{table*}

\subsection{Validation on Temporal-Based Methods}
% \label{sec:temporal}
An intuitive concern is whether \ournet~can still improve temporal‑based 3D object detection since \ournet~has considered semantic consistency in background regions.
To address this concern, we construct a real-world OOD task following~\cite{liu2023bevfusion} where the model is trained on the \emph{non-rainy} data but evaluated on \emph{rainy} and \emph{night} validation data based on BEVDet4D~\cite{huang2022bevdet4d}.
As shown in Table~\ref{tab:nus_real}, \ournet~enhances BEVDet4D not only in rainy but also the night validation scenes, 
which demonstrates the broad applicability of our method.
Note that a video demo is available in our supplementary material.
% It is recommended to check the video demo in our supplementary.

\subsection{Ablation Studies and Visualizations}
To examine \ournet, we provide qualitative results generated by various settings as shown in Figure~\ref{fig:abla}.
Compared with \emph{no velocity adaptation}, applying the foreground preservation loss $\mathcal{L}_{{obj}}$ preserves all annotated objects, while applying the dual-frequency background optimization terms $\mathcal{L}_{{div}}$ and $\mathcal{L}_{{bg}}$ improve editing intensity and enforce semantic consistency.
Eventually, introducing all loss terms achieves the best results. Detailed results of hyper-parameter selection are put in Appendix \emph{\textbf{E}}.
In addition, we provide qualitative visualizations of monocular (top) and multi-view (bottom) object detection as shown in Figure~\ref{fig:vis}. 
% Based on the KITTI dataset, we show that it is easy for \ournet~to augment the original training data into various OOD scenarios since we only require original images and their object annotations (\ie 2D bounding boxes).
% In addition, it is obvious that \ournet~also supports the training data enhancement for multi-view 3D object detection based on the nuScenes dataset.
More visualizations and the results of another powerful flow model, \ie FLUX~\cite{flux2024}, are available in Appendix \emph{\textbf{F}}.
% In Appendix D, we offer more visualizations to demonstrate our effectiveness.
% Meanwhile, we further provide the results of another powerful pre-trained T2I flow model, \ie FLUX~\cite{flux2024}.

\begin{table}[t]
  \setlength\tabcolsep{6pt}
  \fontsize{9pt}{\baselineskip}\selectfont
  \centering

  \begin{tabular}{l|l|ccc}
    \toprule
    % \multirow{2}{*}{Metric} & \multirow{2}{*}{Method} & \multicolumn{3}{c}{nuScenes Real-World} \\
    Metric & Method & Rainy & Night & Avg. \\
    \midrule
    \multirow{2}{*}{mAP} & BEVDet4D & 32.39 & 19.40 & 25.89  \\
    & ~$\bullet~$\ournet~(\textit{3k Snow}) & \textbf{33.65} & \textbf{22.04} & \textbf{27.85}  \\
    \midrule
    \multirow{2}{*}{NDS} & BEVDet4D & 45.06 & 28.29 & 36.68  \\
    & ~$\bullet~$\ournet~(\textit{3k Snow}) & \textbf{45.78} & \textbf{28.78} & \textbf{37.28} \\   
    \bottomrule
  \end{tabular}
    \caption{\label{tab:nus_real} Detection results of the temporal-based method on \emph{real-world} scenarios, regarding mAP and NDS.}
\end{table}

\section{Conclusion}
In this paper, we propose a novel rectified flow adaptation method, namely \ournet, aiming to improve model robustness via training data enhancement in vision-centric 3D object detection.
Specifically, our method performs frequency-based decomposition for the velocity fields of pre-trained T2I flow models. Then, \ournet~devises two strategies:
1) High-frequency foreground preservation aims to maintain all 3D object geometry via a foreground preservation loss.
2) Dual-frequency background optimization introduces the diversity loss to fully exploit pre-trained T2I flow models and the background regularization term to prevent unexpected collapse in background regions.
Experiments on monocular, multi-view and temporal-based multi-view 3D object detection demonstrate the effectiveness of \ournet~in enhancing model robustness.

\clearpage

% \title{Supplementary Material of DriveFlow: Rectified Flow Adaptation for \\ Robust 3D Object Detection in Autonomous Driving}
% \author{ }
% \affiliations{}
% \maketitle
\section{Supplementary Material}
\setcounter{secnumdepth}{1} %May be changed to 1 or 2 if section numbers are desired.
\renewcommand{\thesection}{\Alph{section}}

In this supplementary material, we first provide a comprehensive clarification of the related methodologies. Subsequently, we present additional implementation details, experimental results, analyses, and visualizations of \ournet. The supplementary material is structured as follows:

\begin{itemize}
    \item Appendix~\ref{supp:related} reviews relevant literature on vision-centric 3D object detection.
    \item Appendix~\ref{supp:imple_details} elaborates on the implementation specifics and training procedures of \ournet.
    \item Appendix~\ref{supp:data_details} presents details regarding the construction of the KITTI-C, nuScenes-C, and real-world datasets.
    \item Appendix~\ref{supp:res} provides detailed experimental results.
     \item Appendix~\ref{supp:abla} offers an additional analysis of hyperparameter selection for \ournet. 
     \item Appendix~\ref{supp:vis} demonstrates visualizations generated by \ournet, utilizing SD3 and FLUX.
\end{itemize}

\section{Related Work}
\label{supp:related}
In autonomous driving, vision-centric 3D object detection represents a critical component for environmental perception and scene understanding even in the latest end-to-end methods~\cite{hu2023planning}.
Traditional approaches~\cite{zhou2019objects,ye2020hvnet} rely on LiDAR sensors which achieve precise depth estimation but impose additional hardware costs and complexity.
This limitation has motivated other paradigms shift toward camera-based solutions, \ie monocular and multi-view 3D object detection,  offering cost-effective alternatives while reducing hardware requirements.
For instance, Monocular 3D object detection methods either leverage extra pre-trained depth estimation modules to estimate the depths from a single image~\cite{xu2018multi,zou2021devil}, or generate pseudo-LiDAR~\cite{wang2019pseudo,marethinking,reading2021categorical} to get accurate detections.
As for multi-view 3D object detection, these methods~\cite{chen2017multi,li2022bevformer,huang2022bevdet4d,liu2023bevfusion} significantly improve depth estimation accuracy through multi-perspective geometric understanding and effectively solve occlusion issues, thereby demonstrating better robustness and precision in 3D object detection.

The growing trend of vision-centric perception systems in autonomous vehicles highlights the critical need for efficient yet accurate 3D object detection methods that perform stably across diverse driving scenarios.

\section{Implementation Details}
\label{supp:imple_details}
For the data enhancement part, we implement all methods in PyTorch~\cite{paszke2019pytorch} according to their official repositories.
All experiments are conducted with NVIDIA A100 (80GB of memory) GPUs and every baseline is executed on a single GPU.
Following FlowEdit~\cite{kulikov2024flowedit}, we adopt Stable-Diffusion-3-medium~\cite{esser2024scaling} from the Diffusers library of Hugging Face~\cite{von-platen-etal-2022-diffusers} in the manuscript and set $N_{max}=33$ and $T=50$.
We set the hype-parameters $\lambda_1=5,\lambda_2=1,\lambda_3=1,N_n=5$ in default.
For the Gaussian blur, we set the constant kernel size $k=5$ and $\sigma=1$. The image sizes are set to $1248\times368$ in KITTI and $1344\times768$ in nuScenes.

As for the model training, all 3D detectors are trained on a simple combination of original and augmented data based on their official settings. Specifically, we choose the model with the best performance on the original validation set of KITTI for monocular 3D detectors and then evaluate on KITTI-C.
In addition, we set an equal number of training epochs (\ie 24) for all multi-view 3D detectors following ~\cite{li2022bevformer} on nuScenes and then evaluate on nuScene-C. For details on real-world tasks, please refer to Appendix~\ref{supp:data_details}.

\section{Dataset Details}
\label{supp:data_details}
To evaluate the model robustness of monocular 3D detectors, we adopt the KITTI-C benchmark~\cite{lin2025monotta}, including 13 OOD scenarios based on the original KITTI validation set. 
These OOD scenarios are devised by~\cite{hendrycks2018benchmarking} and we transform the original KITTI validation set into four categories, \ie Noise, Blur, Weather, and Digital.
Similarly, we leverage the nuScenes-C benchmark from the widely used Robo3D benchmark~\cite{xie2025benchmarking} to evaluate the model robustness of multi-view 3D detectors, containing a total of 8 simulated OOD scenarios.

\begin{figure*}[t]
\centering
\includegraphics[width=\textwidth]{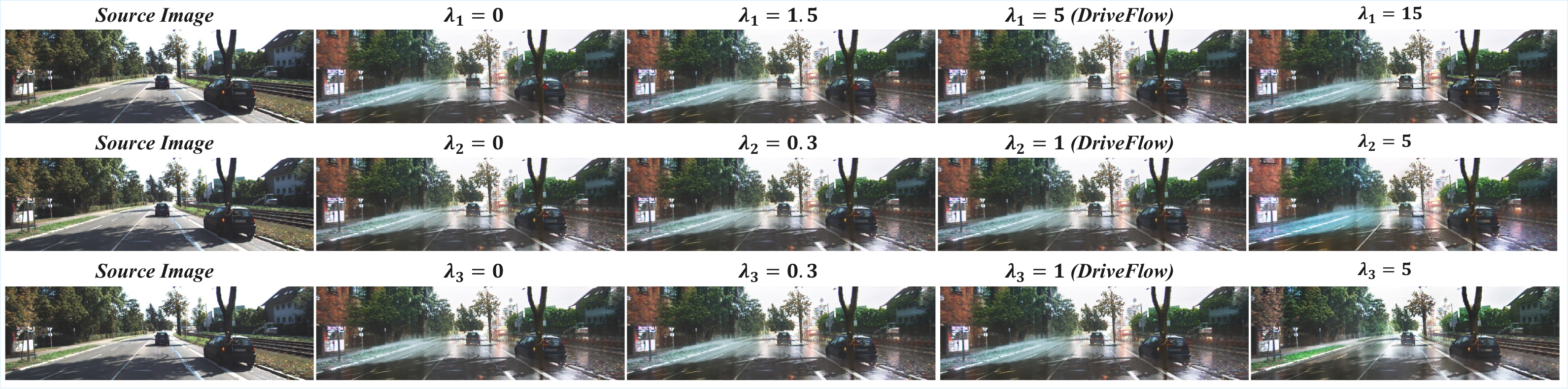}
\caption{Ablation studies on the hyper-parameter selection.}
\label{fig:hyper}
\end{figure*}

\begin{table*}[!h]
  \setlength\tabcolsep{3pt}
  \fontsize{9pt}{\baselineskip}\selectfont
  \centering

  \begin{tabular}{l|l|cccccccccccc}
    \toprule
    \multirow{2}{*}{Metric} & \multirow{2}{*}{Method} & \multicolumn{9}{c}{nuScenes-C} \\
    \cmidrule(lr){3-11} 
    & & Brightness & CameraCrash & ColorQuant & Fog & FrameLost & LowLight & MotionBlur & Snow & Avg. \\
    \midrule
    % \multirow{3}{*}{mAP} & BEVFormer-tiny & 24.26 & 14.89 & 23.91 & 21.98 & 20.43 & 16.11 & 20.78 & 10.83 & 19.15 \\
    % & ~$\bullet$ DriveGEN (\textit{3k Snow}) & 26.04  & 15.79  & 25.78  & 24.13  & 21.13  & 17.20  & 22.39  & 11.72  & 20.52  \\
    % % & ~$\bullet~$ FlowEdit (\textit{3k Snow}) \\
    % & ~$\bullet~$\ournet~(\textit{3k Snow}) & \textbf{26.26} & \textbf{16.76} & \textbf{25.82} & \textbf{24.45} & \textbf{21.72} & \textbf{17.87} & \textbf{22.53 }& \textbf{12.62} & \textbf{21.00} \\
    % % \midrule
    \multirow{3}{*}{NDS} & BEVFormer-tiny & 34.93 & 28.04 & 34.56 & 33.04 & 31.42 & 28.42 & 31.81 & 23.27 & 30.69 \\
    & ~$\bullet$ DriveGEN (\textit{3k Snow}) & 37.14  & 30.02  & \textbf{36.93}  & 35.35  & \textbf{33.32}  & 30.42  & 34.23  & 24.09  & 32.69  \\
    % & ~$\bullet~$ FlowEdit (\textit{3k Snow}) \\
    & ~$\bullet~$\ournet~(\textit{3k Snow}) & \textbf{37.20}  & \textbf{30.64} & 36.52 & \textbf{35.93} & 33.09 & \textbf{30.71} & \textbf{34.28} & \textbf{25.05} & \textbf{32.93}  \\  
    \bottomrule
  \end{tabular}
    \caption{\label{tab:nus-nds} Detection results on nuScenes-C, regarding nuScenes Detection Score.}
\end{table*}

\begin{table*}[!h]
  \setlength\tabcolsep{5pt}
  \fontsize{9pt}{\baselineskip}\selectfont
  \centering
    \begin{tabular}{ll|ll}
    \toprule
        Method & Pre-trained Diffusion Model & Time (s/img) \\
        \midrule
        DriveGEN & Stable Diffusion 1.5  & 176.63 \\
        DriveGEN & Stable Diffusion 2.1 base   & 168.23 \\
        \midrule
        \ournet (Ours) & Stable Diffusion 3.5 medium  & 5.04 ($>30\times$ speed-up) \\
        \ournet (Ours) & FLUX.1-dev & 13.78 ($>12\times$ speed-up) \\
        \bottomrule
    \end{tabular}
    \caption{Comparison on the size of $1248\times368\times3$ images for efficiency exploration.}
    \label{tab:supp-time}
\end{table*}

For practical use, it is essential to evaluate model robustness in real-world OOD scenarios. To this end, we construct a real-world transfer task based on the scene description mentioned in the previous method~\cite{liu2023bevfusion}.
Specifically, we remove all \emph{rainy} training scenes in the training set of nuScenes while validating the detector within \emph{night} (\ie in-distribution but rare) and \emph{rainy} (out-of-distribution) scenes. 
As shown in Table~3 in the manuscript, \ournet~improves the model performance in both cases, demonstrating the effectiveness of the proposed method.

\section{More Experimental Results}
\label{supp:res}
In this section, we aim to provide more detailed experimental results.
We first provide the results regarding nuScenes Detection Score (NDS) of the nuScenes-C dataset as shown in Table~\ref{tab:nus-nds}.
Then, we introduce the detailed results for all categories (\ie Car, Pedestrian and Cyclist) on the KITTI-C dataset as shown in Table~\ref{tab:kitti-c-supp}, \ref{tab:kitti-c-supp-ped} and \ref{tab:kitti-c-supp-cyc}. 

On the one hand, the proposed \ournet~still achieves the best average performance regarding NDS as shown in Table~\ref{tab:nus-nds}, which gives a similar observation to Table~2 in the manuscript.
On the other hand, we provide detailed evaluations for minor classes, specifically Pedestrian (Table~\ref{tab:kitti-c-supp-ped}) and Cyclist (Table~\ref{tab:kitti-c-supp-cyc}).
As emphasized in the main manuscript, autonomous driving datasets inherently exhibit class imbalance, characterized by a substantial dominance of cars over other categories. Consequently, it is crucial to ensure the model maintains robust detection performance across minority classes, rather than overly relying on the dominant class.
Table~\ref{tab:kitti-c-supp-ped} and \ref{tab:kitti-c-supp-cyc} show that \ournet~still achieves the best average performance across diverse OOD scenarios, further demonstrating the effectiveness and contribution of our method.

Then, we further conduct more explorations on efficiency as shown in Table~\ref{tab:supp-time}. 
Obviously, inversion-based approaches suffer from computational inefficiency since reverting to noise maps requires more time.
In contrast, \ournet~leverages pre-trained Text-to-Image (T2I) flow models (\ie Stable Diffusion 3 medium~\cite{esser2024scaling} and FLUX-1.dev~\cite{flux2024}), enabling more powerful and efficient generation.
Furthermore, it is worth emphasizing that the proposed method substantially reduces augmentation time, thereby enabling the practical and scalable deployment of such approaches in real-world applications.

\begin{figure*}[t]
\centering
\includegraphics[width=\textwidth]{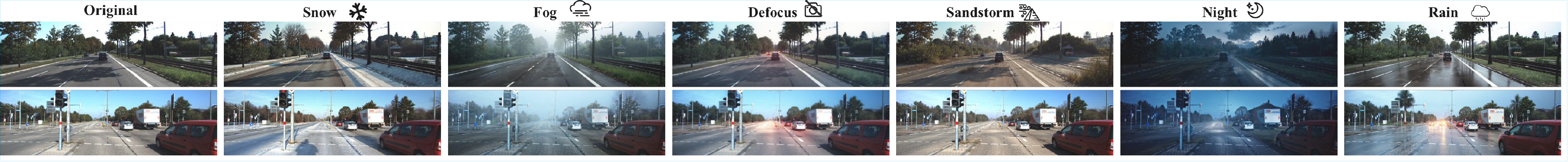}
\caption{Qualitative visualizations of \ournet~based on FLUX within various scenes on KITTI.}
\label{fig:flux}
\end{figure*}

\begin{figure*}[t]
\centering
\includegraphics[width=\textwidth]{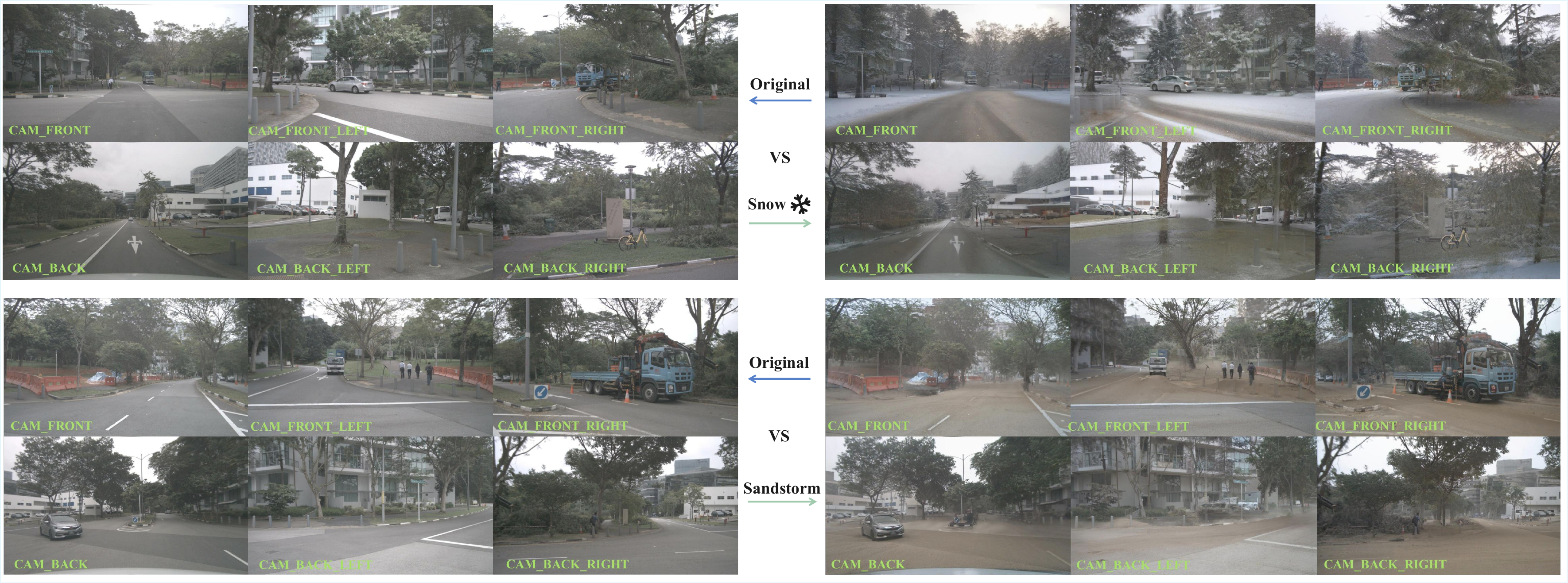}
\caption{Qualitative visualizations of \ournet~within various scenes on nuScenes. Better check \emph{\textbf{the video demo}}.}
\label{fig:supp-vis}
\end{figure*}

\section{More Ablation Studies}
\label{supp:abla}
To examine \ournet, we present additional qualitative results guided by different hyper-parameter settings.
As shown in Figure~\ref{fig:hyper}, if any hyper-parameter is set to zero, we only achieve a sub-optimal result due to the lack of object preservation or background optimization.
As it nears the recommended value, we can get better visualizations which preserve 3D objects and the semantic consistency background with sufficient editing intensity.
Furthermore, even if we set the hyper-parameters to relatively large values, \ournet~still performs well, demonstrating the proposed method is insensitive and stable to hyper-parameters.

\section{More Qualitative Visualizations}
\label{supp:vis}
In this section, we provide more qualitative results generated by FLUX~\cite{flux2024} based on the training images of the KITTI dataset. Then, we also offer more qualitative results based on the training images of the nuScenes dataset enhanced by \ournet.

As shown in Figure~\ref{fig:flux}, \ournet~is also easy to extend to various pre-trained T2I flow models following FlowEdit~\cite{kulikov2024flowedit}.
In addition, as shown in Figure~\ref{fig:supp-vis}, the additional qualitative visualizations further demonstrate the effectiveness of our method.
Moreover, we further provide a video demo of \ournet~in the supplementary material. 

As emphasized in the main paper, \ournet~has considered semantic consistency in background regions and thus \ournet~can still improve temporal‑based 3D object detection by enhancing the training data and maintaining the temporal consistency of augmented scenes.

\begin{table*}[t]
    \begin{center}
    \fontsize{9pt}{\baselineskip}\selectfont
    \setlength\tabcolsep{3pt}

    \begin{tabular}{l|ccc|ccc|cccc|ccc|c}
         \toprule
         \multicolumn{15}{c}{{{Car}, IoU @ 0.7, 0.5, 0.5}} \\
         \midrule
         \multirow{2}{*}{Method}  &
         % \multirow{2}{*}{\shortstack{Inversion-\\free}} &
         % \multirow{2}{*}{\shortstack{Training-\\free}} &
         \multicolumn{3}{c|}{Noise} & 
         \multicolumn{3}{c|}{Blur} & 
         \multicolumn{4}{c|}{Weather} & 
         \multicolumn{3}{c|}{Digital} & 
         \multirow{2}{*}{Avg.} \\
        \cmidrule(lr){2-4} \cmidrule(lr){5-7} \cmidrule(lr){8-11} \cmidrule(lr){12-14} 
        & Gauss. & Shot & Impul. & Defoc. & Glass & Motion & Snow & Frost & Fog & Brit. & Contr. & Pixel & Sat. \\
         \midrule
        Monoflex%~\cite{zhang2021objects}
          & 13.06 &  20.91 &  14.09 &  20.17 &  28.59 &  30.34 &  33.64 &  30.31 &  19.58 &  45.22 &  20.01 &  29.07 &  38.85 & 26.45 \\
        \midrule
        
        ~$\bullet~$ Color Jitter (\textit{Trad.})  & 9.55 & 15.81 & 11.90 & 22.67 & 25.38 & 30.12 & 34.08 & 30.00 & 19.29 & 42.10 & 19.93 & 17.17 & 36.48 & 24.19  \\
        ~$\bullet~$ Brightness (\textit{Trad.})  & 11.44 & 19.42 & 12.73 & 11.18 & 18.95 & 22.03 & 26.64 & 21.70 & 13.04 & 39.61 & 13.08 & 21.11 & 29.73 & 20.05 \\
        
        \midrule
        ~$\bullet~$ ControlNet (\textit{Snow})  & 0.32 & 1.18 & 1.28 & 4.65 & 11.32 & 17.04 & {22.84} & 19.72 & 9.53 & 34.79 & 8.73 & 1.25 & 16.88 & 11.50  \\  
        % ~$\bullet~$ ControlNet (\textit{3 × Aug.})  & 0.70 & 1.14 & 0.29 & 0.35 & 0.37 & 0.83 &  4.07 & 3.05 &  1.20 & 9.43 & 0.94 & 0.48 & 3.20 & 2.00  \\  
        ~$\bullet~$ ControlNet (\textit{6 × Aug.})  & 0.00 & 0.00 & 0.00 &  0.00 & 0.00 & 1.94 &  0.00 & 0.00 &  0.00 & 0.00 & 0.00 & 0.00 & 0.00 & 0.15 \\  
         \midrule
        ~$\bullet~$ Freecontrol (\textit{Snow})  & 20.39 & 28.05 & 22.27 & 12.52 & 21.67 & 21.09 & 27.97 & 17.91 & 10.07 & 35.99 & 9.90 & 28.56 & 35.27 & 22.44 \\  
        % ~$\bullet~$ Freecontrol (\textit{3 × Aug.}) & 15.43 & 21.25 & 14.04 & 15.33 & 19.22 & 16.53 & 22.01 & 17.41 & 15.18 & 24.66 & 16.66 & 25.39 & 31.44 & 19.58 \\  
        ~$\bullet~$ Freecontrol (\textit{6 × Aug.}) & 12.85 & 18.71 & 16.31 & 12.06 & 16.31 & 12.86 & 17.64 & 16.18 & 14.23 & 24.52 & 15.64 & 22.56 & 24.28 & 17.24 \\  
         % \midrule
        ~$\bullet~$ DriveGEN (\textit{Snow}) & 16.48 & 26.72 & 24.98 & 31.17 & 35.55 & 38.13 &  {41.39} & {38.61} & 27.85 & \textbf{49.76} & 29.28 & 38.88 & {44.12} & 34.07  \\  
        % ~$\bullet~$ DriveGEN (\textit{3 × Aug.}) & {25.63} & {37.04} & {29.13} & {34.13} & {39.15} & 36.81 & 38.58 & 37.98 & 33.93 & 45.39 & 34.66 & 39.36 & 43.81 & 36.58  \\  
        ~$\bullet~$ DriveGEN (\textit{6 × Aug.}) & 24.77 & 33.79 & 28.27 & {36.92} &{40.33}& {40.45} & {40.60} & {40.56} & {38.10} & 44.83 & {39.28} & {41.81} & 44.05 & {37.98} \\ 
        \midrule
        ~$\bullet~$ FlowEdit (\textit{Snow}) &  10.50 & 16.12 & 15.52 & 19.48 & 28.25 & 24.18 & 26.81 & 26.36 & 10.39 & 39.26 & 12.06 & 22.47 & 33.20 & 21.89 \\ 
        ~$\bullet~$ \ournet~(\textit{Snow}) &  27.91 & 39.16 & 32.57 & 40.48 & 42.76 & 43.32 & 42.60 & \textbf{43.41} & 40.62 & 47.76 & 42.00 & 44.78 & \textbf{45.45} & 40.99  \\ 
        % ~$\bullet~$ \ournet~(\textit{3 × Aug.}) & \textbf{31.95} & \textbf{41.00} & \textbf{35.18} & \textbf{43.42} & \textbf{43.90} & \textbf{44.09} & \textbf{42.71} & 43.20 & \textbf{45.94} & 48.99 & \textbf{47.15} & \textbf{46.02} & 44.63 & \textbf{42.94} \\ 
        ~$\bullet~$ \ournet~(\textit{6 × Aug.}) & \textbf{31.67} & \textbf{42.54} & \textbf{36.24} & \textbf{44.71} & \textbf{44.48} & \textbf{45.84} & \textbf{42.86} & 43.26 & \textbf{45.16} & 49.12 & \textbf{45.73} & \textbf{46.09} & \textbf{45.82} & \textbf{43.35} \\
        % \midrule
        % \midrule
        %  \multicolumn{15}{c}{{{Pedestrian}, IoU @ 0.7, 0.5, 0.5}} \\
        % \midrule
         \bottomrule
         \end{tabular}
    \end{center}
        \caption{\label{tab:kitti-c-supp} Comparison on {KITTI-C}, severity {level 1} regarding {Mean $AP_{3D|R_{40}}$}. The \textbf{bold} number indicates the best result.
    }
\end{table*}

\begin{table*}[!h]
    \begin{center}
    \fontsize{9pt}{\baselineskip}\selectfont
    \setlength\tabcolsep{3pt}
    \begin{tabular}{l|ccc|ccc|cccc|ccc|c}
         \toprule
         \multicolumn{15}{c}{{{Pedestrian}, IoU @ 0.7, 0.5, 0.5}} \\
         \midrule
         \multirow{2}{*}{Method}  &
         \multicolumn{3}{c|}{Noise} & 
         \multicolumn{3}{c|}{Blur} & 
         \multicolumn{4}{c|}{Weather} & 
         \multicolumn{3}{c|}{Digital} & 
         \multirow{2}{*}{Avg.} \\
        \cmidrule(lr){2-4} \cmidrule(lr){5-7} \cmidrule(lr){8-11} \cmidrule(lr){12-14} 
        & Gauss. & Shot & Impul. & Defoc. & Glass & Motion & Snow & Frost & Fog & Brit. & Contr. & Pixel & Sat. \\
         \midrule
        Monoflex & 1.16 & 3.54 & 0.78 & 8.06 & 17.70 & 15.26 & 12.72 & 9.25 & 5.61 & 19.87 & 5.35 & 1.49 & 8.65 & 8.42  \\
        \midrule

        ~$\bullet~$ Color Jitter (\textit{Trad.})  & 0.99 & 3.53 & 1.41 & 10.05 & 14.63 & 12.00 & 14.73 & 12.13 & 7.72 & 19.02 & 9.23 & 1.11 & 11.29 & 9.07 \\
        ~$\bullet~$ Brightness (\textit{Trad.})  & 0.63 & 1.85 & 1.20 & 5.01 & 13.72 & 12.15 & 7.93 & 5.84 & 1.87 & 16.89 & 2.30 & 0.48 & 3.77 & 5.67 \\
        
        \midrule
        ~$\bullet~$ ControlNet (\textit{Snow})  & 0.00 & 0.00 & 0.00 & 1.78 & 8.32 & 5.84 & 4.03 & 3.75 & 1.31 & 11.00 & 1.23 & 0.00 & 0.93 & 2.94 \\
        ~$\bullet~$ ControlNet (\textit{6 × Aug.})  & 0.00 & 0.00 & 0.00 & 0.00 & 0.36 & 0.00 & 0.00 & 0.00 & 0.00 & 2.50 & 0.00 & 0.00 & 0.00 & 0.22 \\ 
         \midrule
        ~$\bullet~$ Freecontrol (\textit{Snow})  & 3.62 & 5.17 & 3.21 & 3.23 & 5.16 & 5.94 & 4.68 & 3.52 & 4.28 & 7.70 & 4.22 & 8.51 & 7.70 & 5.15 \\
        ~$\bullet~$ Freecontrol (\textit{6 × Aug.}) & 5.99 & 9.10 & 9.08 & 4.80 & 7.36 & 5.84 & 7.22 & 7.58 & 7.10 & 10.81 & 9.40 & 9.90 & 10.24 & 8.03 \\  
         % \midrule
        ~$\bullet~$ DriveGEN (\textit{Snow}) & 1.25 & 3.22 & 4.56 & 16.21 & 19.88 & 20.61 & 19.80 & 14.46 & 8.80 & 24.62 & 9.11 & 9.77 & 18.04 & 13.10 \\
        ~$\bullet~$ DriveGEN (\textit{6 × Aug.}) & 6.34 & 9.22 & 7.71 & 17.60 & 19.21 & 20.20 & 16.88 & 16.19 & 15.72 & 23.57 & 16.75 & 14.37 & 17.31 & 15.47 \\
        \midrule
        ~$\bullet~$ FlowEdit (\textit{Snow}) & 1.56 & 2.18 & 1.75 & 9.28 & 15.53 & 13.11 & 6.58 & 4.62 & 1.15 & 15.14 & 1.35 & 4.99 & 9.15 & 6.65 \\
        ~$\bullet~$ \ournet~(\textit{Snow}) & 12.05 & 17.12 & 13.54 & 13.04 & 15.39 & 19.05 & 17.32 & 17.29 & 19.70 & 23.42 & 21.21 & 17.01 & 18.40 & 17.27 \\
        
       ~$\bullet~$ \ournet~(\textit{6 × Aug.}) & \textbf{18.14} & \textbf{22.37} & \textbf{16.03} & \textbf{22.25} & \textbf{23.08} & \textbf{24.45} & \textbf{20.69} & \textbf{22.81} & \textbf{20.23} & \textbf{25.67} & \textbf{21.99} & \textbf{21.66} & \textbf{22.75} & \textbf{21.70} \\ 

        \midrule
        \midrule

        MonoGround & 2.67 & 3.25 & 5.76 & 17.57 & 18.91 & 17.71 & 12.96 & 9.35 & 4.37 & 24.15 & 5.89 & 3.27 & 7.16 & 10.23 \\

        \midrule
        
        ~$\bullet~$ Color Jitter (\textit{Trad.})  & 2.44 & 3.24 & 4.11 & 15.37 & 18.46 & 16.50 & 15.45 & 12.38 & 9.14 & 24.71 & 9.82 & 2.07 & 7.81 & 10.89 \\
        ~$\bullet~$ Brightness (\textit{Trad.})  & 2.79 & 4.10 & 7.61 & 14.51 & 14.13 & 14.52 & 12.12 & 12.66 & 5.55 & 20.80 & 5.36 & 2.36 & 10.31 & 9.75 \\
        
        \midrule
        ~$\bullet~$ ControlNet (\textit{Snow})  & 1.85 & 1.03 & 0.81 & 7.18 & 9.97 & 8.32 & 1.32 & 2.92 & 1.37 & 12.99 & 1.40 & 0.28 & 0.90 & 3.87 \\
        ~$\bullet~$ ControlNet (\textit{6 × Aug.})  & 0.00 & 0.00 & 0.00 & 0.00 & 0.00 & 1.67 & 0.00 & 0.00 & 0.00 & 0.00 & 0.00 & 0.00 & 0.00 & 0.13 \\ 
         \midrule
        ~$\bullet~$ Freecontrol (\textit{Snow})  & 10.04 & 13.09 & 11.70 & 14.91 & 12.61 & 13.03 & 15.14 & 12.35 & 8.13 & 15.69 & 12.34 & 16.32 & 14.57 & 13.07 \\
        ~$\bullet~$ Freecontrol (\textit{6 × Aug.}) & 6.81 & 7.54 & 7.67 & 5.54 & 5.45 & 5.10 & 2.19 & 1.31 & 3.63 & 8.59 & 5.35 & 8.86 & 6.26 & 5.72 \\  
         % \midrule
        ~$\bullet~$ DriveGEN (\textit{Snow}) & 6.22 & 6.89 & 9.27 & 15.03 & 17.24 & 18.81 & 16.40 & 13.38 & 9.81 & 23.28 & 11.50 & 12.04 & 13.92 & 13.37 \\
        ~$\bullet~$ DriveGEN (\textit{6 × Aug.}) & 9.70 & 13.68 & 13.52 & 17.00 & 17.74 & 20.39 & 17.23 & 18.79 & 15.50 & 23.09 & 15.80 & 20.99 & 18.34 & 17.06 \\
        \midrule
        ~$\bullet~$ FlowEdit (\textit{Snow}) & 2.27 & 3.46 & 2.41 & 11.78 & 13.68 & 13.90 & 10.50 & 7.97 & 9.53 & 20.90 & 10.94 & 13.05 & 10.71 & 10.08 \\
        ~$\bullet~$ \ournet~(\textit{Snow}) & 14.22 & 17.44 & 10.97 & 21.04 & \textbf{21.81} & \textbf{22.91} & 19.55 & 18.92 & 20.00 & \textbf{27.25} & \textbf{23.66} & 18.17 & \textbf{22.63} & 19.89 \\
        
        ~$\bullet~$ \ournet~(\textit{6 × Aug.}) & \textbf{15.29} & \textbf{18.15} & \textbf{12.90} & \textbf{21.67} & {21.78} & {22.87} & \textbf{19.95} & \textbf{21.73} & \textbf{21.16} & {26.17} & {23.22} & \textbf{22.47} & {21.77} & \textbf{20.70} \\
        
         \bottomrule
         \end{tabular}
    \end{center}
        \caption{\label{tab:kitti-c-supp-ped} Comparison on {KITTI-C} for the Pedestrian category regarding {Mean $AP_{3D|R_{40}}$}.
    }
\end{table*}

\begin{table*}[t]
    \begin{center}
    \fontsize{9pt}{\baselineskip}\selectfont
    \setlength\tabcolsep{3pt}
    \begin{tabular}{l|ccc|ccc|cccc|ccc|c}
         \toprule
         \multicolumn{15}{c}{{{Cyclist}, IoU @ 0.7, 0.5, 0.5}} \\
         \midrule
         \multirow{2}{*}{Method}  &
         \multicolumn{3}{c|}{Noise} & 
         \multicolumn{3}{c|}{Blur} & 
         \multicolumn{4}{c|}{Weather} & 
         \multicolumn{3}{c|}{Digital} & 
         \multirow{2}{*}{Avg.} \\
        \cmidrule(lr){2-4} \cmidrule(lr){5-7} \cmidrule(lr){8-11} \cmidrule(lr){12-14} 
        & Gauss. & Shot & Impul. & Defoc. & Glass & Motion & Snow & Frost & Fog & Brit. & Contr. & Pixel & Sat. \\
         \midrule
        Monoflex & 0.43 & 2.41 & 0.64 & 2.76 & 8.30 & 9.14 & \textbf{12.85} & 11.09 & 5.73 & \textbf{17.44} & 4.84 & 3.25 & 9.89 & 6.83 \\

        \midrule
        
        ~$\bullet~$ Color Jitter (\textit{Trad.})  & 0.63 & 3.15 & 1.91 & 1.62 & 3.43 & 7.92 & 11.03 & 10.09 & 4.60 & 12.41 & 4.61 & 1.43 & 10.23 & 5.62 \\
        ~$\bullet~$ Brightness (\textit{Trad.})  & 0.21 & 1.16 & 0.25 & 1.33 & 3.45 & 6.14 & 9.67 & 8.81 & 4.89 & 13.66 & 5.82 & 2.02 & 7.93 & 5.03 \\
        
        \midrule
        ~$\bullet~$ ControlNet (\textit{Snow})  & 0.00 & 0.30 & 0.00 & 0.00 & 3.77 & 4.29 & 7.27 & 6.47 & 6.97 & 15.79 & 6.49 & 1.67 & 2.54 & 4.27 \\
        ~$\bullet~$ ControlNet (\textit{6 × Aug.})  & 0.00 & 0.00 & 0.00 & 0.00 & 0.00 & 0.00 & 0.00 & 0.00 & 0.00 & 0.00 & 0.00 & 0.00 & 0.00 & 0.00 \\ 
         \midrule
        ~$\bullet~$ Freecontrol (\textit{Snow})  & 1.58 & 4.43 & 1.72 & 0.00 & 0.39 & 0.94 & 3.97 & 1.52 & 0.54 & 5.26 & 0.68 & 1.07 & 7.50 & 2.28 \\
        ~$\bullet~$ Freecontrol (\textit{6 × Aug.}) & 0.19 & 0.24 & 0.45 & 0.00 & 0.31 & 0.00 & 0.55 & 0.52 & 1.01 & 2.13 & 2.12 & 0.61 & 0.81 & 0.69 \\  
         % \midrule
        ~$\bullet~$ DriveGEN (\textit{Snow}) & 0.70 & 1.27 & 0.61 & 1.34 & 5.26 & 5.27 & 10.90 & 7.12 & 3.73 & 15.14 & 4.37 & 1.74 & 11.24 & 5.28 \\
        ~$\bullet~$ DriveGEN (\textit{6 × Aug.}) & 0.53 & 0.93 & 0.54 & 3.07 & 10.95 & 9.38 & 11.12 & 12.60 & 9.07 & 15.39 & 10.81 & 1.99 & 8.05 & 7.26 \\
        \midrule
        ~$\bullet~$ FlowEdit (\textit{Snow}) & 0.28 & 0.27 & 0.43 & 1.46 & 5.46 & 4.11 & 3.55 & 3.05 & 0.60 & 9.92 & 0.93 & 3.16 & 5.13 & 2.95 \\
        ~$\bullet~$ \ournet~(\textit{Snow}) & 2.15 & 8.43 & 5.94 & 6.57 & 8.22 & 11.16 & 12.46 & 13.08 & 8.09 & 16.66 & 8.68 & 14.54 & 14.79 & 10.06 \\
        
        ~$\bullet~$ \ournet~(\textit{6 × Aug.}) & \textbf{6.38} & \textbf{12.97} & \textbf{8.73} & \textbf{14.87} & \textbf{16.57} & \textbf{17.03} & 11.67 & \textbf{15.35} & \textbf{14.02} & 17.40 & \textbf{15.16} & \textbf{15.81} & \textbf{18.30} & \textbf{14.17} \\ 

        \midrule
        \midrule

        MonoGround & 0.21 & 1.86 & 1.34 & 0.83 & 2.93 & 2.23 & 5.00 & 3.43 & 0.94 & 11.48 & 1.21 & 2.04 & 5.92 & 3.03 \\

        \midrule
        
        ~$\bullet~$ Color Jitter (\textit{Trad.})  & 0.39 & 2.67 & 2.11 & 0.31 & 2.03 & 2.19 & 5.38 & 4.63 & 1.12 & \textbf{13.64} & 1.67 & 2.89 & 5.00 & 3.39 \\
        ~$\bullet~$ Brightness (\textit{Trad.})  & 0.06 & 0.61 & 0.22 & 0.36 & 1.33 & 1.06 & 4.72 & 2.32 & 1.41 & 6.87 & 0.78 & 0.90 & 2.81 & 1.80 \\
        
        \midrule
        ~$\bullet~$ ControlNet (\textit{Snow})  & 0.00 & 0.00 & 0.52 & 0.00 & 0.77 & 1.33 & 0.44 & 1.10 & 0.14 & 6.77 & 0.30 & 0.50 & 0.54 & 0.95 \\
        ~$\bullet~$ ControlNet (\textit{6 × Aug.})  & 0.00 & 0.00 & 0.00 & 0.00 & 0.00 & 0.00 & 0.00 & 0.00 & 0.00 & 0.00 & 0.00 & 0.00 & 0.00 & 0.00 \\ 
         \midrule
        ~$\bullet~$ Freecontrol (\textit{Snow})  & 0.46 & 0.70 & 0.32 & 1.07 & 0.17 & 0.50 & 1.60 & 0.91 & 0.21 & 4.48 & 0.17 & 1.93 & 4.50 & 1.31 \\
        ~$\bullet~$ Freecontrol (\textit{6 × Aug.}) & 0.00 & 0.34 & 0.62 & 0.00 & 0.42 & 0.56 & 0.00 & 0.00 & 1.00 & 2.07 & 0.83 & 1.25 & 0.74 & 0.60 \\  
         % \midrule
        ~$\bullet~$ DriveGEN (\textit{Snow}) & 0.13 & 0.81 & 0.38 & 0.31 & 2.23 & 3.66 & 3.96 & 2.02 & 0.90 & 8.46 & 1.74 & 2.07 & 4.58 & 2.40 \\
        ~$\bullet~$ DriveGEN (\textit{6 × Aug.}) & 1.49 & 2.16 & 1.66 & 3.30 & 5.97 & 5.55 & 5.64 & 5.49 & 2.49 & 9.37 & 3.48 & 3.79 & 5.65 & 4.31 \\
        \midrule
        ~$\bullet~$ FlowEdit (\textit{Snow}) & 0.62 & 1.22 & 0.57 & 0.75 & 2.22 & 3.63 & 2.28 & 5.02 & 3.30 & 5.72 & 3.48 & 2.79 & 2.19 & 2.60 \\
        ~$\bullet~$ \ournet~(\textit{Snow}) & 1.40 & 3.69 & \textbf{4.39} & 3.56 & 6.51 & \textbf{8.02} & 8.67 & 9.14 & 8.05 & 12.01 & \textbf{9.45} & 9.92 & 10.43 & 7.33 \\
        
        ~$\bullet~$ \ournet~(\textit{6 × Aug.}) & \textbf{3.75} & \textbf{4.49} & 3.51 & \textbf{3.99} & \textbf{6.87} & 7.66 & \textbf{9.73} & \textbf{9.35} & \textbf{8.33} & 11.53 & 8.54 & \textbf{11.76} & \textbf{10.35} & \textbf{7.68} \\
        
         \bottomrule
         \end{tabular}
    \end{center}
        \caption{\label{tab:kitti-c-supp-cyc} Comparison on {KITTI-C} for the Cyclist category regarding {Mean $AP_{3D|R_{40}}$}.
    }
\end{table*}

% \clearpage

\subsubsection{Acknowledgments.}
This work was supported by NSFC with Grant No. 62573371, by the Basic Research Project No. HZQB-KCZYZ-2021067 of Hetao Shenzhen-HK S\&T Cooperation Zone, by Guangdong S\&T Programme with Grant No. 2024B0101030002, by the Shenzhen General Program No. JCYJ20220530143600001, by the Shenzhen Outstanding Talents Training Fund 202002, by the Guangdong Research Project No.2017ZT07X152 and No. 2019CX01X104, by the Guangdong Provincial Key Laboratory of Future Networks of Intelligence (Grant No. 2022B1212010001), by the Guangdong Provincial Key Laboratory of BigData Computing CHUK-Shenzhen, by the NSFC 61931024\&12326610\&62293482, by the Key Area R\&D Program of Guangdong Province with grant No. 2018B030338001, by the Shenzhen Key Laboratory of Big Data and Artificial Intelligence (Grant No. SYSPG20241211173853027), by China Association for Science and Technology Youth Care Program, by the Shenzhen-Hong Kong Joint Funding No. SGDX20211123112401002, and by Tencent \& Huawei Open Fund.

\bibliography{aaai2026}

@String(ICCV= {Int. Conf. Comput. Vis.})

@String(AAAI = {AAAI})

@String(ICCV  = {ICCV})

@inproceedings{zhang2021objects,
  title={Objects are different: Flexible monocular 3d object detection},
  author={Zhang, Yunpeng and Lu, Jiwen and Zhou, Jie},
  booktitle={Proceedings of the IEEE/CVF Conference on Computer Vision and Pattern Recognition},
  pages={3289--3298},
  year={2021}
}

@inproceedings{qin2022monoground,
  title={Monoground: Detecting monocular 3d objects from the ground},
  author={Qin, Zequn and Li, Xi},
  booktitle={Proceedings of the IEEE/CVF Conference on Computer Vision and Pattern Recognition},
  pages={3793--3802},
  year={2022}
}

@inproceedings{xu2023mononerd,
  title={Mononerd: Nerf-like representations for monocular 3d object detection},
  author={Xu, Junkai and Peng, Liang and Cheng, Haoran and Li, Hao and Qian, Wei and Li, Ke and Wang, Wenxiao and Cai, Deng},
  booktitle={Proceedings of the IEEE/CVF International Conference on Computer Vision},
  pages={6814--6824},
  year={2023}
}

@inproceedings{geiger2012we,
  title={Are we ready for autonomous driving? the kitti vision benchmark suite},
  author={Geiger, Andreas and Lenz, Philip and Urtasun, Raquel},
  booktitle={Proceedings of the IEEE/CVF Conference on Computer Vision and Pattern Recognition},
  pages={3354--3361},
  year={2012},
  organization={IEEE}
}

@inproceedings{chen2017multi,
  title={Multi-view 3d object detection network for autonomous driving},
  author={Chen, Xiaozhi and Ma, Huimin and Wan, Ji and Li, Bo and Xia, Tian},
  booktitle={Proceedings of the IEEE conference on Computer Vision and Pattern Recognition},
  pages={1907--1915},
  year={2017}
}

@inproceedings{wang2019pseudo,
  title={Pseudo-lidar from visual depth estimation: Bridging the gap in 3d object detection for autonomous driving},
  author={Wang, Yan and Chao, Wei-Lun and Garg, Divyansh and et. al.},
  booktitle={Proceedings of the IEEE/CVF Conference on Computer Vision and Pattern Recognition},
  pages={8445--8453},
  year={2019}
}

@inproceedings{chen2023voxelnext,
  title={Voxelnext: Fully sparse voxelnet for 3d object detection and tracking},
  author={Chen, Yukang and Liu, Jianhui and Zhang, Xiangyu and Qi, Xiaojuan and Jia, Jiaya},
  booktitle={Proceedings of the IEEE/CVF Conference on Computer Vision and Pattern Recognition},
  pages={21674--21683},
  year={2023}
}

@inproceedings{caesar2020nuscenes,
  title={nuscenes: A multimodal dataset for autonomous driving},
  author={Caesar, Holger and Bankiti, Varun and Lang, Alex H and et.al.},
  booktitle={Proceedings of the IEEE/CVF conference on computer vision and pattern recognition},
  pages={11621--11631},
  year={2020}
}

@inproceedings{sun2020scalability,
  title={Scalability in perception for autonomous driving: Waymo open dataset},
  author={Sun, Pei and Kretzschmar, Henrik and Dotiwalla, Xerxes and Chouard, Aurelien and Patnaik, Vijaysai and Tsui, Paul and Guo, James and Zhou, Yin and Chai, Yuning and Caine, Benjamin and others},
  booktitle={Proceedings of the IEEE/CVF conference on computer vision and pattern recognition},
  pages={2446--2454},
  year={2020}
}

@inproceedings{lin2025monotta,
  title={MonoTTA: Fully Test-Time Adaptation for Monocular 3D Object Detection},
  author={Lin, Hongbin and Zhang, Yifan and Niu, Shuaicheng and Cui, Shuguang and Li, Zhen},
  booktitle={European Conference on Computer Vision},
  pages={96--114},
  year={2025},
  organization={Springer}
}

@inproceedings{oh2025monowad,
  title={MonoWAD: Weather-Adaptive Diffusion Model for Robust Monocular 3D Object Detection},
  author={Oh, Youngmin and Kim, Hyung-Il and Kim, Seong Tae and Kim, Jung Uk},
  booktitle={European Conference on Computer Vision},
  year={2025}
}

@inproceedings{li2022bevformer,
  title={Bevformer: Learning bird’s-eye-view representation from multi-camera images via spatiotemporal transformers},
  author={Li, Zhiqi and Wang, Wenhai and Li, Hongyang and Xie, Enze and Sima, Chonghao and Lu, Tong and Qiao, Yu and Dai, Jifeng},
  booktitle={European conference on computer vision},
  pages={1--18},
  year={2022},
  organization={Springer}
}

@inproceedings{zhang2023adding,
  title={Adding conditional control to text-to-image diffusion models},
  author={Zhang, Lvmin and Rao, Anyi and Agrawala, Maneesh},
  booktitle={Proceedings of the IEEE/CVF International Conference on Computer Vision},
  pages={3836--3847},
  year={2023}
}

@inproceedings{tumanyan2023plug,
  title={Plug-and-play diffusion features for text-driven image-to-image translation},
  author={Tumanyan, Narek and Geyer, Michal and Bagon, Shai and Dekel, Tali},
  booktitle={Proceedings of the IEEE/CVF Conference on Computer Vision and Pattern Recognition},
  pages={1921--1930},
  year={2023}
}

@inproceedings{mo2024freecontrol,
  title={Freecontrol: Training-free spatial control of any text-to-image diffusion model with any condition},
  author={Mo, Sicheng and Mu, Fangzhou and Lin, Kuan Heng and Liu, Yanli and Guan, Bochen and Li, Yin and Zhou, Bolei},
  booktitle={Proceedings of the IEEE/CVF Conference on Computer Vision and Pattern Recognition},
  pages={7465--7475},
  year={2024}
}

@inproceedings{rombach2022high,
  title={High-resolution image synthesis with latent diffusion models},
  author={Rombach, Robin and Blattmann, Andreas and Lorenz, Dominik and Esser, Patrick and Ommer, Bj{\"o}rn},
  booktitle={Proceedings of the IEEE/CVF conference on computer vision and pattern recognition},
  pages={10684--10695},
  year={2022}
}

@article{song2020denoising,
title={Denoising diffusion implicit models},
author={Song, Jiaming and Meng, Chenlin and Ermon, Stefano},   journal={arXiv preprint arXiv:2010.02502},
year={2020} 
}

@article{ravi2024sam,
  title={Sam 2: Segment anything in images and videos},
  author={Ravi, Nikhila and Gabeur, Valentin and Hu, Yuan-Ting and Hu, Ronghang and Ryali, Chaitanya and Ma, Tengyu and Khedr, Haitham and R{\"a}dle, Roman and Rolland, Chloe and Gustafson, Laura and others},
  journal={arXiv preprint arXiv:2408.00714},
  year={2024}
}

@inproceedings{chen2024internvl,
  title={Internvl: Scaling up vision foundation models and aligning for generic visual-linguistic tasks},
  author={Chen, Zhe and Wu, Jiannan and Wang, Wenhai and Su, Weijie and Chen, Guo and Xing, Sen and Zhong, Muyan and Zhang, Qinglong and Zhu, Xizhou and Lu, Lewei and others},
  booktitle={Proceedings of the IEEE/CVF Conference on Computer Vision and Pattern Recognition},
  pages={24185--24198},
  year={2024}
}

@inproceedings{hendrycks2018benchmarking,
  title={Benchmarking Neural Network Robustness to Common Corruptions and Perturbations},
  author={Hendrycks, Dan and Dietterich, Thomas},
  booktitle={International Conference on Learning Representations},
  year={2018}
}

@inproceedings{liu2023bevfusion,
  title={Bevfusion: Multi-task multi-sensor fusion with unified bird's-eye view representation},
  author={Liu, Zhijian and Tang, Haotian and Amini, Alexander and Yang, Xinyu and Mao, Huizi and Rus, Daniela L and Han, Song},
  booktitle={2023 IEEE international conference on robotics and automation (ICRA)},
  pages={2774--2781},
  year={2023},
  organization={IEEE}
}

@inproceedings{paszke2019pytorch,
  title={Pytorch: An imperative style, high-performance deep learning library},
  author={Paszke, Adam and Gross, Sam and Massa, Francisco and Lerer, Adam and Bradbury, James and Chanan, Gregory and Killeen, Trevor and Lin, Zeming and Gimelshein, Natalia and Antiga, Luca and others},
  booktitle={Advances in neural information processing systems},
  volume={32},
  year={2019}
}

@inproceedings{yan2024monocd,
  title={MonoCD: Monocular 3D Object Detection with Complementary Depths},
  author={Yan, Longfei and Yan, Pei and Xiong, Shengzhou and Xiang, Xuanyu and Tan, Yihua},
  booktitle={Proceedings of the IEEE/CVF Conference on Computer Vision and Pattern Recognition},
  pages={10248--10257},
  year={2024}
}

@article{ramesh2022hierarchical,
  title={Hierarchical Text-Conditional Image Generation with CLIP Latents},
  author={Ramesh, Aditya and Dhariwal, Prafulla and Nichol, Alex and Chu, Casey and Chen, Mark},
  journal={arXiv preprint arXiv:2204.06125},
  year={2022}
}

@article{zhao2023uni,
  title={Uni-ControlNet: All-in-One Control to Text-to-Image Diffusion Models},
  author={Zhao, Shihao and Chen, Dongdong and Chen, Yen-Chun and Bao, Jianmin and Zhang, Dongdong and Yuan, Lu and Zhang, Han and Li, Dong and Chen, Baining and Zhang, Lu},
  journal={arXiv preprint arXiv:2305.16322},
  year={2023}
}

@article{qin2023unicontrol,
  title={UniControl: A Unified Diffusion Model for Controllable Visual Generation In the Wild},
  author={Qin, Yuheng and Zhang, Zhaoyang and Zhang, Han and Li, Dong and Chen, Baining and Zhang, Lu and Gu, Jiuxiang and Zhang, Dongdong},
  journal={arXiv preprint arXiv:2305.11147},
  year={2023}
}

@article{sun2022panacea,
  title={Panacea: A framework for diverse, controllable generation of urban scenes},
  author={Sun, Lin and Cao, Liangliang and Zheng, Yang and Wang, Bo and Li, Zhiming and Xie, Wenzhe and Sun, Xu},
  journal={arXiv preprint arXiv:2207.10701},
  year={2022}
}

@article{hong2021magicdrive,
  title={MagicDrive: 3D Controllable Driving Video Synthesis Using Layout and 3D Representations},
  author={Hong, Yuming and Xu, Weiyao and Yang, Yi and Zhou, Hao and Wang, Xiaoxue},
  journal={Proceedings of the IEEE/CVF International Conference on Computer Vision (ICCV)},
  year={2021}
}

@inproceedings{lin2025drivegen,
  title={Drivegen: Generalized and robust 3d detection in driving via controllable text-to-image diffusion generation},
  author={Lin, Hongbin and Guo, Zilu and Zhang, Yifan and Niu, Shuaicheng and Li, Yafeng and Zhang, Ruimao and Cui, Shuguang and Li, Zhen},
  booktitle={Proceedings of the Computer Vision and Pattern Recognition Conference},
  pages={27497--27507},
  year={2025}
}

@inproceedings{pu2025monodgp,
  title={Monodgp: Monocular 3D object detection with decoupled-query and geometry-error priors},
  author={Pu, Fanqi and Wang, Yifan and Deng, Jiru and Yang, Wenming},
  booktitle={Proceedings of the Computer Vision and Pattern Recognition Conference},
  pages={6520--6530},
  year={2025}
}

@inproceedings{wang2023exploring,
  title={Exploring object-centric temporal modeling for efficient multi-view 3d object detection},
  author={Wang, Shihao and Liu, Yingfei and Wang, Tiancai and Li, Ying and Zhang, Xiangyu},
  booktitle={Proceedings of the IEEE/CVF international conference on computer vision},
  pages={3621--3631},
  year={2023}
}

@article{wang2020tent,
  title={Tent: Fully test-time adaptation by entropy minimization},
  author={Wang, Dequan and Shelhamer, Evan and Liu, Shaoteng and Olshausen, Bruno and Darrell, Trevor},
  journal={arXiv preprint arXiv:2006.10726},
  year={2020}
}

@article{kulikov2024flowedit,
  title={Flowedit: Inversion-free text-based editing using pre-trained flow models},
  author={Kulikov, Vladimir and Kleiner, Matan and Huberman-Spiegelglas, Inbar and Michaeli, Tomer},
  journal={arXiv preprint arXiv:2412.08629},
  year={2024}
}

@article{wang2024taming,
  title={Taming rectified flow for inversion and editing},
  author={Wang, Jiangshan and Pu, Junfu and Qi, Zhongang and Guo, Jiayi and Ma, Yue and Huang, Nisha and Chen, Yuxin and Li, Xiu and Shan, Ying},
  journal={arXiv preprint arXiv:2411.04746},
  year={2024}
}

@inproceedings{zhang2025geobev,
  title={Geobev: Learning geometric bev representation for multi-view 3d object detection},
  author={Zhang, Jinqing and Zhang, Yanan and Qi, Yunlong and Fu, Zehua and Liu, Qingjie and Wang, Yunhong},
  booktitle={Proceedings of the AAAI Conference on Artificial Intelligence},
  volume={39},
  pages={9960--9968},
  year={2025}
}

@inproceedings{li2025rctrans,
  title={Rctrans: Radar-camera transformer via radar densifier and sequential decoder for 3d object detection},
  author={Li, Yiheng and Yang, Yang and Lei, Zhen},
  booktitle={Proceedings of the AAAI Conference on Artificial Intelligence},
  volume={39},
  pages={5048--5056},
  year={2025}
}

@article{bai2025qwen2,
  title={Qwen2. 5-vl technical report},
  author={Bai, Shuai and Chen, Keqin and Liu, Xuejing and Wang, Jialin and Ge, Wenbin and Song, Sibo and Dang, Kai and Wang, Peng and Wang, Shijie and Tang, Jun and others},
  journal={arXiv preprint arXiv:2502.13923},
  year={2025}
}

@article{russell2025gaia,
  title={Gaia-2: A controllable multi-view generative world model for autonomous driving},
  author={Russell, Lloyd and Hu, Anthony and Bertoni, Lorenzo and Fedoseev, George and Shotton, Jamie and Arani, Elahe and Corrado, Gianluca},
  journal={arXiv preprint arXiv:2503.20523},
  year={2025}
}

@article{hu2023gaia,
  title={Gaia-1: A generative world model for autonomous driving},
  author={Hu, Anthony and Russell, Lloyd and Yeo, Hudson and Murez, Zak and Fedoseev, George and Kendall, Alex and Shotton, Jamie and Corrado, Gianluca},
  journal={arXiv preprint arXiv:2309.17080},
  year={2023}
}

@misc{flux2024,
    author={Black Forest Labs},
    title={FLUX},
    year={2024},
    howpublished={\url{https://github.com/black-forest-labs/flux}},
}

@inproceedings{zheng2023layoutdiffusion,
  title={Layoutdiffusion: Controllable diffusion model for layout-to-image generation},
  author={Zheng, Guangcong and Zhou, Xianpan and Li, Xuewei and Qi, Zhongang and Shan, Ying and Li, Xi},
  booktitle={Proceedings of the IEEE/CVF Conference on Computer Vision and Pattern Recognition},
  pages={22490--22499},
  year={2023}
}

@article{liu2022flow,
  title={Flow straight and fast: Learning to generate and transfer data with rectified flow},
  author={Liu, Xingchao and Gong, Chengyue and Liu, Qiang},
  journal={arXiv preprint arXiv:2209.03003},
  year={2022}
}

@inproceedings{esser2024scaling,
  title={Scaling rectified flow transformers for high-resolution image synthesis},
  author={Esser, Patrick and Kulal, Sumith and Blattmann, Andreas and Entezari, Rahim and M{\"u}ller, Jonas and Saini, Harry and Levi, Yam and Lorenz, Dominik and Sauer, Axel and Boesel, Frederic and others},
  booktitle={Forty-first international conference on machine learning},
  year={2024}
}

@article{huang2022bevdet4d,
  title={Bevdet4d: Exploit temporal cues in multi-camera 3d object detection},
  author={Huang, Junjie and Huang, Guan},
  journal={arXiv preprint arXiv:2203.17054},
  year={2022}
}

@article{xie2025benchmarking,
  title={Benchmarking and Improving Bird's Eye View Perception Robustness in Autonomous Driving},
  author={Xie, Shaoyuan and Kong, Lingdong and Zhang, Wenwei and Ren, Jiawei and Pan, Liang and Chen, Kai and Liu, Ziwei},
  journal={IEEE Transactions on Pattern Analysis and Machine Intelligence},
  year={2025},
  publisher={IEEE}
}

@inproceedings{hu2023planning,
  title={Planning-oriented autonomous driving},
  author={Hu, Yihan and Yang, Jiazhi and Chen, Li and Li, Keyu and Sima, Chonghao and Zhu, Xizhou and Chai, Siqi and Du, Senyao and Lin, Tianwei and Wang, Wenhai and others},
  booktitle={Proceedings of the IEEE/CVF conference on computer vision and pattern recognition},
  pages={17853--17862},
  year={2023}
}

@inproceedings{ye2020hvnet,
  title={Hvnet: Hybrid voxel network for lidar based 3d object detection},
  author={Ye, Maosheng and Xu, Shuangjie and Cao, Tongyi},
  booktitle={Proceedings of the IEEE/CVF conference on computer vision and pattern recognition},
  pages={1631--1640},
  year={2020}
}

@article{zhou2019objects,
  title={Objects as points},
  author={Zhou, Xingyi and Wang, Dequan and Kr{\"a}henb{\"u}hl, Philipp},
  journal={arXiv preprint arXiv:1904.07850},
  year={2019}
}

@inproceedings{xu2018multi,
  title={Multi-level fusion based 3d object detection from monocular images},
  author={Xu, Bin and Chen, Zhenzhong},
  booktitle={Proceedings of the IEEE conference on computer vision and pattern recognition},
  pages={2345--2353},
  year={2018}
}

@inproceedings{zou2021devil,
  title={The devil is in the task: Exploiting reciprocal appearance-localization features for monocular 3d object detection},
  author={Zou, Zhikang and Ye, Xiaoqing and Du, Liang and Cheng, Xianhui and Tan, Xiao and Zhang, Li and Feng, Jianfeng and Xue, Xiangyang and Ding, Errui},
  booktitle={Proceedings of the IEEE/CVF International Conference on Computer Vision},
  pages={2713--2722},
  year={2021}
}

@inproceedings{marethinking,
  title={Rethinking Pseudo-LiDAR Representation},
  author={Ma, Xinzhu and Liu, Shinan and Xia, Zhiyi and Zhang, Hongwen and Zeng, Xingyu and Ouyang, Wanli},
  booktitle={European Conference on Computer Vision},
  year={2020}
}

@inproceedings{reading2021categorical,
  title={Categorical depth distribution network for monocular 3d object detection},
  author={Reading, Cody and Harakeh, Ali and Chae, Julia and Waslander, Steven L},
  booktitle={Proceedings of the IEEE/CVF Conference on Computer Vision and Pattern Recognition},
  pages={8555--8564},
  year={2021}
}

@misc{von-platen-etal-2022-diffusers,
  author = {Patrick von Platen and Suraj Patil and Anton Lozhkov and Pedro Cuenca and Nathan Lambert and Kashif Rasul and Mishig Davaadorj and Dhruv Nair and Sayak Paul and William Berman and Yiyi Xu and Steven Liu and Thomas Wolf},
  title = {Diffusers: State-of-the-art diffusion models},
  year = {2022},
  publisher = {GitHub},
  journal = {GitHub repository},
  howpublished = {\url{https://github.com/huggingface/diffusers}}
}

\end{document}